\documentclass{article} %
\usepackage{iclr2020_conference,times}
\pdfoutput=1

\renewcommand{\headrulewidth}{0pt}
\iclrfinalcopy

\usepackage{amsmath}
\usepackage{amsfonts}
\usepackage{bm}

\def\eqref#1{equation~\ref{#1}}

\def\1{\bm{1}}

\DeclareMathAlphabet{\mathsfit}{\encodingdefault}{\sfdefault}{m}{sl}
\SetMathAlphabet{\mathsfit}{bold}{\encodingdefault}{\sfdefault}{bx}{n}

\usepackage{graphicx}
\usepackage{hyperref}
\usepackage{url}
\usepackage{amsmath}
\usepackage{amssymb}
\usepackage{amsthm}
\usepackage{comment}
\usepackage{listings}
\usepackage{float}

\usepackage{subcaption}

\usepackage{appendix}

\lstdefinestyle{mystyle}{
    captionpos=b,                    
    numbers=left,          
    xleftmargin=.25\textwidth
}
\newtheorem{theorem}{Theorem}
\newtheorem{corollary}{Corollary}
\usepackage{dcolumn}

\newcolumntype{C}{>{\centering}p{1cm}}
\newcolumntype{E}{D{.}{.}{5.0}}

\newcommand{\parheader}[1]{\noindent {\bf #1}}

\definecolor{mred}{rgb}{.80,.12,.30}
\definecolor{grey}{rgb}{0.5,0.5,0.5}
\definecolor{purple2}{rgb}{.75,0,.85}
\definecolor{pistachio}{rgb}{0.58, 0.77, 0.45}
\definecolor{steelblue}{rgb}{.10,.40,.85}

\newcommand{\C}[0]{\mathcal{S}}

\newcommand{\NComplexInvariants}[0]{12}

\definecolor{mygreen}{HTML}{467E7E}
\definecolor{mygray}{rgb}{0.5,0.5,0.5}
\definecolor{myblue}{HTML}{144B7D}
\definecolor{myorange}{HTML}{B25A00}

\lstdefinestyle{myC}{
  language=C,
  backgroundcolor=\color{white},
  basicstyle=\linespread{0.9}\ttfamily\small,
  breakatwhitespace=false,
  breaklines=true,
  commentstyle=[\bfseries\color{mygreen},    %
  deletekeywords={...},
  escapeinside={<@}{@>},
  extendedchars=true,
  keepspaces=true,
  keywordstyle=\bfseries\color{myblue!70}, 
  otherkeywords={uint},     
  deletekeywords={get,angle, gamma, invariant},
  emph = { certiq_prove, match, end},
  emphstyle=\bfseries\color{myorange!70},
  showspaces=false, 
  showstringspaces=false,
  showtabs=false, 
  stringstyle=\color{mymauve},
  tabsize=1,
 morecomment=[f][\bfseries\color{mymauve!70}][0]{@},
 morecomment=[f][\bfseries\color{mygreen}][0]{//},
}

\usepackage{booktabs}
\usepackage{siunitx}
\newcommand{\specialcellbold}[2][b]{%
  \bfseries
  \sisetup{text-rm=\bfseries}%
  \begin{tabular}[#1]{@{}c@{}}#2\end{tabular}%
}

\usepackage{enumitem}
\usepackage{authblk}
\title{CLN2INV: Learning Loop Invariants with \\ Continuous Logic Networks}

\newcommand*\samethanks[1][\value{footnote}]{\footnotemark[#1]}
\author{Gabriel Ryan\thanks{Co-student leads listed in alphabetical order; each contributed equally.} ,  Justin Wong\samethanks{} ,  Jianan Yao\samethanks{} ,   Ronghui Gu,   Suman Jana \\
Columbia University\\
\texttt{\{gabe,suman\}@cs.columbia.edu}\\
\texttt{\{justin.wong,jy3022,ronghui.gu\}@columbia.edu} \\
}

\begin{document}

\maketitle

\begin{abstract}
Program verification offers a framework for ensuring program correctness and therefore systematically eliminating different classes of bugs. Inferring loop invariants is one of the main challenges behind automated verification of real-world programs which often contain many loops. In this paper, we present Continuous Logic Network (CLN), a novel neural architecture for automatically learning loop invariants directly from program execution traces. Unlike existing neural networks, CLNs can learn precise and explicit representations of formulas in Satisfiability Modulo Theories (SMT)  for loop invariants from program execution traces. We develop a new sound and complete semantic mapping for assigning SMT formulas to continuous truth values that allows CLNs to be trained efficiently. We use CLNs to implement a new inference system for loop invariants, CLN2INV, that significantly outperforms existing approaches on the popular Code2Inv dataset. CLN2INV is the first tool to solve all 124 theoretically solvable problems in the Code2Inv dataset. Moreover, CLN2INV takes only 1.1 seconds on average for each problem, which is 40$\times$ faster than existing approaches. We further demonstrate that CLN2INV can even learn \NComplexInvariants{} significantly more complex loop invariants than the ones required for the Code2Inv dataset.
\end{abstract}

\section{Introduction}
\label{sec:intro}

Program verification offers a principled approach for systematically eliminating different classes of bugs and proving the correctness of programs. However, as programs have become increasingly complex, real-world program verification often requires prohibitively expensive manual effort~\citep{wilcox2015verdi, gu2016certikos, Chajed2019verifying}. Recent efforts have focused on automating the program verification process, but automated verification of general programs with unbounded loops remains an open problem  \citep{nelson2017hyperkernel, nelson2019scaling}. 

Verifying programs with loops requires determining 
\emph{loop invariants}, which captures the effect of the loop on the program state irrespective of the actual number of loop iterations. Automatically inferring correct loop invariants is a challenging problem that is undecidable in general and difficult to solve in practice \citep{blass2001inadequacy, furia2014loop}. Existing approaches use stochastic search~\citep{sharma2016invariant}, heurstics-based search~\citep{galeotti2015inferring}, PAC learning based on counter examples ~\citep{loopinvgen2017}, or reinforcement learning~\citep{si2018learning}. However, these approaches often struggle to learn complex, real-world loop invariants.

In this paper, we introduce a new approach to learning loop invariants by modeling the loop behavior from program execution traces using a new type of neural architecture. We note that inferring loop invariants can be posed as learning formulas in Satisfiability Modulo Theories (SMT) \citep{handbookofSAT}  over program variables collected from program execution traces
~\citep{nguyen2017counterexample}. 
In principle, Neural networks seem well suited to this task because they can act as universal function approximators and have been successfully applied in various domains that require  modeling of arbitrary functions \citep{hornik1989multilayer, goodfellow2016deep}. However, loop invariants must be represented as explicit SMT formulas to be usable for program verification. Unfortunately, existing methods for extracting logical rules from general neural architectures lack sufficient precision \citep{Augasta2012RuleEF}, while inductive logic learning lacks sufficient expressiveness for use in verification \citep{deepmind2017dilp}.

We address this issue by developing a novel neural architecture, Continuous Logic Network (CLN), which is able to efficiently learn explicit and precise representations of SMT formulas by using continuous truth values. Unlike existing neural architectures, CLNs can represent a learned SMT formula explicitly in its structure and thus allow us to precisely extract the exact formula from a trained model.

In order to train CLNs, we introduce a new semantic mapping for SMT formulas to continuous truth values. Our semantic mapping builds on BL, or basic fuzzy logic \citep{hajek2013metamathematics}, to support general SMT formulas in a continuous logic setting.
We further prove that our semantic model is sound (i.e., truth assignments for the formulas are consistent with their discrete counterparts) and complete (i.e., all formulas can be represented) with regard to the discrete SMT formula space. These properties allow CLNs to represent any quantifier-free SMT formula operating on mixed integer-real arithmetic as an end-to-end differentiable series of operations.

We use CLNs to implement a new inference system for loop invariants, CLN2INV, that significantly outperforms state-of-the-art tools on the Code2Inv dataset by solving all 124 theoretically solvable problems in the dataset. This is 20 problems more than LoopInvGen, the winner of the SyGus 2018 competition loop invariant track \citep{si2018learning}. Moreover, CLN2INV finds invariants for each program in 1.1 seconds
on average, more than 40 times faster than LoopInvGen. We also demonstrate CLN2INV is able to learn complex, real-world loop invariants with combinations of conjunctions and disjunctions of multivariable constraints.

Our main contributions are:
\begin{itemize}[topsep=2pt]
    \item We introduce a new semantic mapping for assigning continuous truth values to SMT formulas that is theoretically grounded and enables learning formulas through backpropagation. We further prove that our semantic model is sound and complete. 
    \item We develop a novel neural architecture, Continuous Logic Networks (CLNs), that to the best of our knowledge is the first to efficiently learn precise and explicit SMT formulas by construction.
    \item We use CLNs to implement a new loop invariant inference system, CLN2INV, that is the first to solve all 124 theoretically solvable problems in the Code2Inv dataset, 20 more than the existing methods. CLN2INV is able to find invariants for each problem in 1.1 seconds on average, 40$\times$ faster than existing systems.
    \item We further show CLN2INV is able to learn \NComplexInvariants{} more complex loop invariants than the ones present in the Code2Inv dataset with combinations of multivariable constraints.
\end{itemize}

\parheader{Related Work.} Traditionally, loop invariant learning has relied on stochastic  or heuristics-guided search \citep{sharma2016invariant,galeotti2015inferring}. Other approaches like  Guess-And-Check analyze traces and discover conjunctions of equalities by solving a system of linear equations ~\citep{sharma2013data}. NumInv extends this approach to identify octagonal bounds from counterexamples in a limited range  ~\citep{nguyen2017counterexample}. LoopInvGen uses PAC learning of CNF using counter-examples \citep{loopinvgen2017}. By contrast, Code2Inv learns to guess loop invariants using reinforcement learning with recurrent and graph neural networks \citep{si2018learning}. However, these approaches struggle to learn complex invariants.
Unlike these works, CLN2INV can efficiently learn complex invariants directly from execution traces.

There is a long line of work on PAC learning of boolean formulas, but learning precise formulas require a prohibitively large number of samples \citep{kearns1994introduction}.  Several recent works use different forms of differentiable logic to learn boolean logic formulas from noisy data \citep{kimmig2012short, deepmind2017dilp, payani2019inductive} or improving adversarial robustness by applying logical rules to training \citep{fischer2019dl2}. 
By contrast, our work learns precise SMT formulas directly by construction, allowing us to learn richer predicates with compact representation in a noiseless setting.

\section{Background}
\label{sec:background}
In this section, we introduce the problem of inferring loop invariants and provide a brief overview of Satisfiability Modulo Theories (SMT), which are used to represent loop invariants. We then provide an introduction to basic fuzzy logic, which we later extend to formulate our new continuous semantic mapping for SMT.

\subsection{Loop Invariants}
\label{sec:back_invariants}

 Loop invariants capture loop behavior irrespective of number of iterations, which is crucial for verifying programs with loops. Given a loop, \texttt{while($C$)\{$S$\}}, a precondition $P$, and a post-condition $Q$, the verification task
involves finding a loop invariant $I$ that can be concluded from the pre-condition and implies the post-condition~\citep{hoare1969axiomatic}. Formally, it must satisfy the following three conditions, in which the second is a Hoare triple describing the loop:
\vspace{2pt}
\begin{align*}
P \implies I &&  \{I \land C \}\ S\ \{I\}  && I \land \lnot C \implies Q
\end{align*}

\parheader{Example of Loop Invariant.} Consider the example loop in Fig.\ref{fig:example_loop}. For a loop invariant to be usable, it must be valid for the precondition $(t=10 \land u=0)$, the recursion step when $t\neq 0$, and the post condition $(u = 20)$ when the loop condition  is no longer satisfied, i.e., $t = 0$.  The correct and precise invariant $I$  for the program is $(2t+u=20)$. 
\begin{center}
\begin{figure}[h]
\hspace{-20pt}
\begin{subfigure}[b]{0.35\textwidth}
\label{fig:P1}
\begin{lstlisting}[language =C, numbers=none, style=myC]
//pre: t=10 /\ u=0
while (t != 0){
  t = t - 1;
  u = u + 2;
}
//post: u=20
\end{lstlisting}
\caption{\label{fig:P1} Example loop}
\end{subfigure}
\qquad
\begin{subfigure}[b]{0.60\textwidth}
The desired loop invariant $I$ for the left program
is a boolean \\ 
function over program variables $t,u$ such that:
\[\forall t\ u,\left\{
\begin{array}{rll}
    t=10 \land u = 0 &\implies I(t,u) & (pre)  \\
    I(t,u) \land (t\neq0) &\implies I(t-1,u+2) & (inv) \\
    I(t,u) \land (t =0) &\implies u=20 & (post) 
\end{array}\right.\]
\caption{The desired and precise loop invariant $I$ is $(2t+u=20)$.}
\end{subfigure}
\setlength{\abovecaptionskip}{15pt}
\setlength{\belowcaptionskip}{-10pt}
\caption{\label{fig:example_loop}Example Loop Invariant inference problem.}
\end{figure}
\end{center}

\subsection{Satisfiability Modulo Theories}

Satisfiability Modulo Theories (SMT) are an extension of Boolean Satisfiability that allow solvers to reason about complex problems efficiently. Loop invariants and other formulas in program verification are usually encoded with quantifier-free SMT. 
A formula $F$ in quantifier-free SMT
can be inductively defined as below:
\begin{align*}
    F := E_1 \bowtie E_2\ |\ \neg F \ | \ F_1 \wedge F_2
\ | \ F_1 \vee F_2 &      & \bowtie \in \{=, \neq, <, >, \leq, \geq\}
\end{align*}
where $E_1$ and $E_2$ are  expressions
of terms.  The loop invariant $(2t+u=20)$ in Fig.~\ref{fig:example_loop}
is an SMT formula. Nonlinear arithmetic theories admit higher-order terms such as $t^2$ and $t*u$, allowing them to express more complex constraints. 
For example, $(\neg(2 \geq t^2))$ is an SMT
formula that is true %
when the value of the high-order term $t^2$ is larger than 2.

\subsection{Basic Fuzzy Logic (BL)}
\label{sec:bl_logic}

Basic fuzzy logic (BL) is a class of logic that uses continuous truth values in the range $[0,1]$ and is  differentiable almost everywhere\footnote{Almost everywhere indicates the function is differentiable everywhere except for a set of measure 0. For example, a Rectified Linear Unit is differentiable almost everywhere except at zero.} \citep{hajek2013metamathematics}. BL defines logical conjunction with functions called {\it t-norms}, which must satisfy specific conditions to ensure that the behavior of the logic is consistent with boolean First Order Logic.
Formally, a t-norm (denoted $\otimes$) in BL is a continuous binary operator over  continuous truth values
satisfying the following conditions: 

\begin{itemize}[topsep=2pt]

\item[] 1) \textit{commutativity} and \textit{associativity}: the order in which a set of t-norms on continuous truth values are evaluated should not change the result.
\begin{center}
$x \otimes y = y \otimes x\ \ \ \ x \otimes (y \otimes z) = (x \otimes y) \otimes z$.
\end{center}
\item[] 2) \textit{monotonicity}: increasing any input value to a t-norm operation should not cause the result to decrease.
\begin{center}
    $x_1 \leq x_2 \implies x_1 \otimes y \leq x_2 \otimes y $
    \end{center}

\item[] 3) \textit{consistency}: the result of any t-norm applied to a truth value and 1 should be 1, and the result of any truth value and 0 should be 0.
    \begin{center}
    $   1 \otimes x = x $. $0 \otimes x = 0$
    \end{center}
\end{itemize}%

\section{Continuous Satisfiability Modulo Theories}
\label{sec:logic}

We introduce a \emph{continuous semantic mapping}, $\C$, for SMT on BL that is end-to-end differentiable. 
The mapping $\C$ associates SMT formulas with continuous truth values
while preserving each formula's semantics. In this paper, we only consider quantifier-free formulas. This process is analogous to constructing t-norms for BL, where a t-norm operates on continuous logical inputs.

We define three desirable properties for continuous semantic mapping $\C$ that will preserve formula semantics while facilitating parameter training with gradient descent:
\begin{enumerate}[topsep=2pt]
    \item $\C(F)$ should be consistent with BL. 
    For any two formulas $F$ and $F'$, where $F(x)$ is satisfied and $F'(x)$ is unsatisfied
    with an assignment $x$ of formula terms, we should have $\C(F')(x) < \C(F)(x)$.
    This will ensure the semantics of SMT formulas are preserved. 
    \item $\C(F)$ should be differentiable almost everywhere. This will facilitate training with gradient descent through backpropogation.
    \item  $\C(F)$ should be increasing everywhere as the terms in the formula approach constraint satisfaction, and decreasing everywhere as the terms in the formula approach constraint violation.
    This ensures there is always a nonzero gradient for training.
\end{enumerate}

 \parheader{Continuous semantic mapping.} 
  We first define the mapping for ``$>$'' (greater-than) and 
 ``$\geq$'' (greater-than-or-equal-to) as well as adopting definitions  for ``$\neg$'', ``$\land$'',
and ``$\lor$'' from BL.
All other operators can be derived from these. 
For example, ``$\leq$'' (less-than-or-equal-to) is derived using  ``$\geq$'' and ``$\neg$'', while ``$=$'' (equality) is then defined as the conjunction of formulas using  ``$\leq$'' and ``$\geq$.''
 Given constants $B>0$ and $\epsilon>0$, we first define the the mapping $\C$ on ``$>$'' and ``$\geq$''
 using shifted and scaled sigmoid functions:
\begin{align*}
    &\C(t > u)  \triangleq \frac{1}{1 + e^{-B(t-u-\epsilon)}} & &\C(t \geq u)  \triangleq \frac{1}{1 + e^{-B(t-u+\epsilon)}}
\end{align*}
\begin{figure}[b]
\centering
\begin{subfigure}[b]{.44\textwidth}
    \centering
    \includegraphics[width=\textwidth]{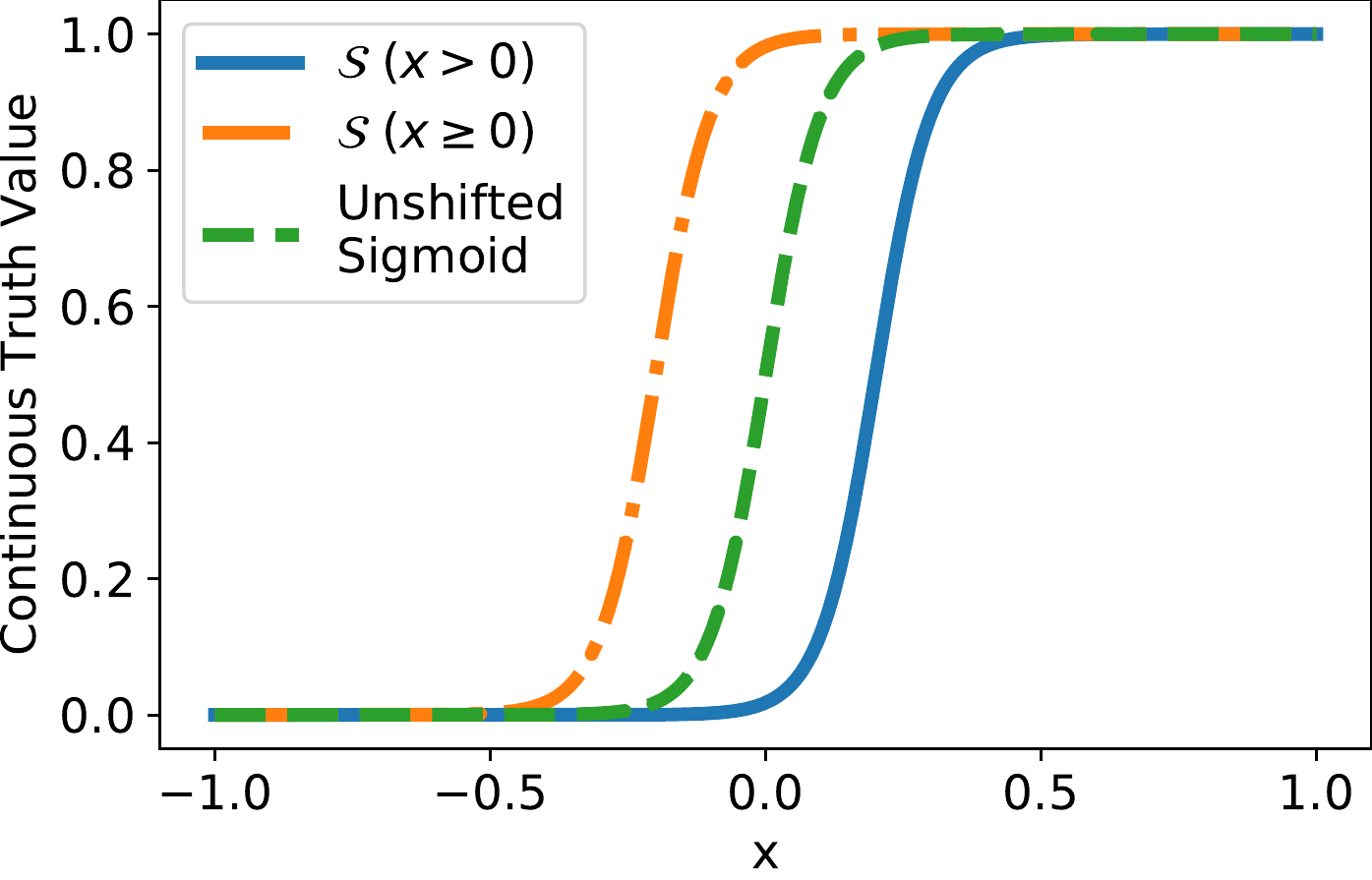}
    \caption{Plot of $\C(x \geq 0)$, $\C(x > 0)$ with sigmoid}
    \label{fig:sigmoids}
\end{subfigure}
\qquad
\begin{subfigure}[b]{.44\textwidth}
    \centering
    \includegraphics[width=\textwidth]{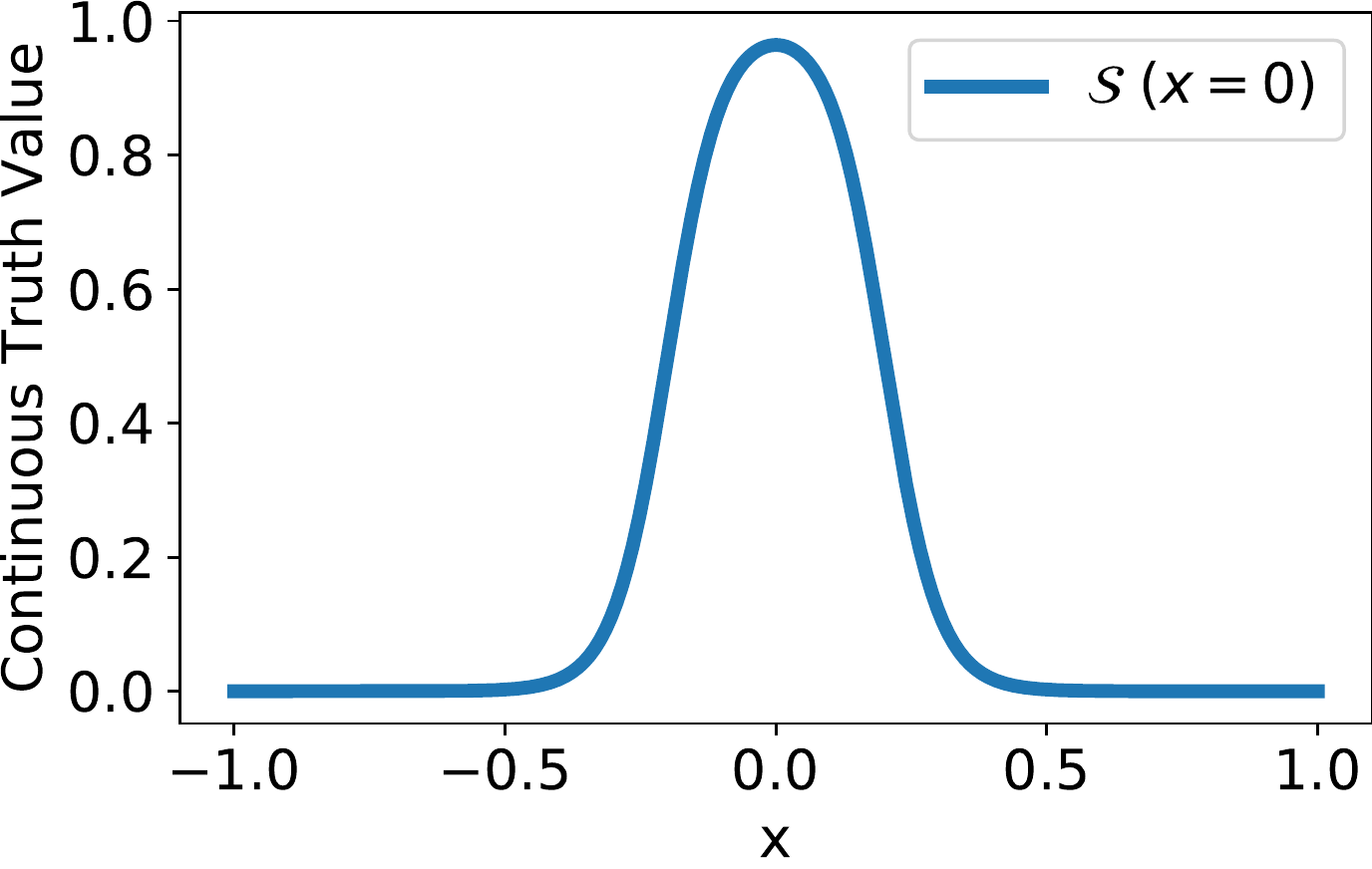}
    \caption{Plot of $\C(x=0)$ with product t-norm}
    \label{fig:sigmoid-equality}
\end{subfigure}
\setlength{\belowcaptionskip}{-10pt}
\caption{\label{fig:ex_predicates}Illustration of the mapping $\C$  on $>,\geq,=$ when $B=20$ and $\epsilon=0.2$}
\end{figure}
Illustrations of shifted sigmoids for $\C(>)$, $\C(\geq)$, and $\C(=)$ are given in Figure \ref{fig:ex_predicates}. The validity of our semantic mapping lie in the following facts, which can be proven with basic algebra.
\begin{align*}
    \lim_{\substack{\epsilon \rightarrow 0^+ \\ B \cdot \epsilon \rightarrow +\infty}} \frac{1}{1 + e^{-B(t-u-\epsilon)}} &= \left\{\begin{array}{lr}
    1 & t > u \\
    0 & t \leq u
\end{array}\right.&
\lim_{\substack{\epsilon \rightarrow 0^+ \\ B \cdot \epsilon \rightarrow +\infty}} \frac{1}{1 + e^{-B(t-u+\epsilon)}} &= \left\{\begin{array}{lr}
    1 & t \geq u \\
    0 & t < u
\end{array}\right.
\end{align*}
When $\epsilon$ goes to zero and $B*\epsilon$ goes to infinity, our continuous mapping
of ``$>$'' and ``$\geq$'' will preserve their
original semantics. Under these conditions, our mapping satisfies all three desirable properties. In practice, for small $\epsilon$ and large $B$, the properties are also satisfied if $|t-u| > \epsilon$.

Next we define the mapping $\C$ for boolean operators ``$\land$'', ``$\lor$'' and ``$\neg$'' using BL. Recall that in BL, a t-norm is a continuous function that behaves like logical conjunction. In \S\ref{sec:bl_logic}, we outlined requirements for a valid t-norm. Three widely used t-norms that satisfy the requirements are the Lukaseiwicz t-norm~\citep{lukasiewicz1930untersuchungen}, the Godel t-norm~\citep{baaz1996infinite}, and the product t-norm~\citep{hajek1996complete}. Each t-norm has a {\it t-conorm} associated with it (denoted $\oplus$), which can be considered as logical disjunction. Given a t-norm $\otimes$, the t-conorm can be derived with DeMorgan's law: $t \oplus u \triangleq \neg(\neg t \otimes \neg u)$.
\begin{align*}
    & &    &\textrm{Lukasiewicz}: & &\textrm{Godel:}  & &\textrm{Product:}\\
    \textrm{t-norm}(\otimes)& &   &max (0,t+u-1) & &min(t,u)  & & t * u\\
    \textrm{t-conorm} (\oplus)& &   &min(t+u, 1)& &max(t, u)& & t + u - t * u
\end{align*}
Given a specific t-norm $\otimes$ and its corresponding t-conorm $\oplus$, it is straightforward to define mappings of ``$\land$'', ``$\lor$'' and ``$\neg$'': 
\begin{align*}
&\C(F_1 \land F_2)  \triangleq  \C(F_1) \otimes \C(F_2) & \C(F_1 \lor F_2) \triangleq \C(F_1) \oplus \C(F_2)&
&\C(\neg F)  \triangleq 1 - \C(F)
\end{align*}
Based on the above definitions, the mapping for  other operators can be derived as follows:
\begin{align*}
    \C(t < u) & = \C(\neg(t \geq u))= \frac{1}{1 + e^{B(t-u+\epsilon)}} & 
    \;\;\; \C(t \leq u) & = \C(\neg (t > u)) = \frac{1}{1 + e^{B(t-u-\epsilon)}} 
\end{align*}
\begin{align*}
    \C(t = u) & = \C((t \geq u) \land (t \leq u)) = \frac{1}{1+e^{-B(t-u+\epsilon)}} \otimes \frac{1}{1+e^{B(t-u-\epsilon)}} 
\end{align*}
The mapping $\C$ on ``$=$'' is valid since the following limit holds (see Appendix \ref{app:limit-equality} for the proof). 
\[
\lim_{\substack{\epsilon \rightarrow 0^+ \\ B \cdot \epsilon \rightarrow +\infty}}\!\!\!\! \C(t = u) 
\;\; =
\lim_{\substack{\epsilon \rightarrow 0^+ \\ B \cdot \epsilon \rightarrow +\infty}}\!\! \frac{1}{1+e^{-B(t-u+\epsilon)}} \otimes \frac{1}{1+e^{B(t-u-\epsilon)}} = \left\{\begin{array}{lr}
1 & t = u \\
0 & t \neq u
\end{array}\right.\]
The mapping for other operators shares similar behavior in the limit, and also fulfill our desired properties under the same conditions.

Using our semantic mapping $\C$, most of the standard operations  of integer and real arithmetic, including addition, subtraction, multiplication, division, and exponentiation, can be used normally and mapped to continuous truth values while keeping the entire formula differentiable. Moreover, any expression in SMT that has an integer or real-valued result can be mapped to continuous logical values via these formulas, although end-to-end differentiability may not be maintained in cases where specific operations are nondifferentiable.

\section{Continuous Logic Networks}
\label{sec:methodology}

In this section, we describe the construction of Continuous Logic Networks (CLNs) based on our continuous semantic mapping for SMT on BL.

\parheader{CLN Construction.} CLNs use our continuous semantic mapping to represent SMT formulas as an end-to-end differentiable model that can learn selected constants in the formula. When constructing a CLN, we work from an {\it SMT Formula Template}, in which every value is marked as either an input term, a constant, or a learnable parameter. Given an SMT Formula Template, we dynamically construct a CLN  as a computational graph, where input terms are marked as model inputs. The operations in each SMT clause are recursively added to the graph, followed by logical operations on clauses. Figure \ref{fig:arch} shows an example formula template and the constructed CLN. We denote the CLN model constructed from the formula template $\C(F)$ as $M_F$.

\parheader{CLN Training.} Once the CLN has been constructed based on a formula template, it is trained with the following optimization. Given a CLN model $M$ constructed from an SMT template %
with learnable parameters $\mathbf{W}$, and  a set $\mathbf{X}$ of valid assignments for the terms in the SMT template, the expected value of the CLN is maximized by minimizing a loss function $\mathcal{L}$ that penalizes model outputs that are less than one. 
A minimum scaling factor $\beta$ is selected, and a hinge loss is applied to the scaling factors ($B$) to force the differentiable predicates to approach sharp cutoffs. The offset $\epsilon$ is also regularized to ensure precision. The overall optimization is formulated as:

\vspace{-5pt}
\begin{align*}
    \max_{\{\mathbf{W}, B, \epsilon\}} \mathbb{E}[M(\mathbf{X}; \mathbf{W}, B, \epsilon)] =  \min_{\{\mathbf{W}, B, \epsilon\}} \sum_{\bm{x} \in \mathbf{X}} \mathcal{L}(M(\bm{x}; \mathbf{W}, B, \epsilon)) + \lambda \sum_{B \in \mathbf{B}} \mathcal{L}_{hinge}(\beta, B) + \gamma ||\epsilon||_2
\end{align*}
\vspace{-5pt}

where $\lambda$ and $\gamma$ are hyperparameters respectively governing the weight assigned to the scaling factor and offset regularization. $\mathcal{L}_{hinge}(\beta, B)$ is defined as $max(0, \beta - B)$, and $\mathcal{L}$ is any loss function strictly decreasing in domain $[0,1]$.

Given a CLN that has been trained to a loss approaching 0 on a given set of valid assignments, we now show that the resulting continuous SMT formula learned by the CLN is consistent with an equivalent boolean SMT formula. In particular, we prove that continuous SMT formulas learned with CLNs are {\it sound} and {\it complete} with regard to SMT formulas on discrete logic. We further prove that a subset of SMT formulas are guaranteed to converge to a globally optimal solution.

\noindent {\bf Soundness.} Given the SMT formula $F$, the CLN model $M_F$ constructed from $\C(F)$ always preserves the truth value of $F$. It indicates that given a valid assignment to the terms $\bm{x}$ in $F$, $F(\bm{x}) = True \iff M_F(\bm{x})=1$ and $F(\bm{x}) = False \iff M_F(\bm{x})=0$.

\noindent {\bf Completeness.} For any SMT formula $F$, a CLN model $M$ can be constructed representing that formula. In other words, CLNs can express all SMT formulas on integers and reals. 

We formally state these properties in Theorem \ref{th:sound_complete} and provide a proof by induction on the constructor in the Appendix \ref{app:proof1}. Before that we need to define a property for t-norms. 

\parheader{Property 1.} $\forall t\ u, (t>0) \text{ and } (u>0) \text{ implies } (t \otimes u > 0)$. 

The product t-norm and Godel t-norm have this property, while the Lukasiewicz t-norm does not.

\begin{theorem}
\label{th:sound_complete}
For any SMT formula $F$, there exists a CLN model $M$, such that
\[\forall \bm{x}\ B\ \epsilon, \ 0 \leq M(\bm{x};B,\epsilon) \leq 1\]
\begin{align*}
F(\bm{x})=True \iff \mkern-10mu \lim_{\substack{\epsilon \rightarrow 0^+\\B\cdot\epsilon\rightarrow\infty}} \mkern-10mu M(\bm{x};B,\epsilon)=1& & & &
F(\bm{x})=False \iff \mkern-10mu \lim_{\substack{\epsilon \rightarrow 0^+\\B\cdot\epsilon\rightarrow\infty}} \mkern-10mu M(\bm{x};B,\epsilon)=0
\end{align*}
as long as the t-norm used in building $M$ satisfies Property 1.
\end{theorem}

\noindent {\bf Optimality.} For a subset of SMT formulas (conjunctions of multiple linear equalities), CLNs are guaranteed to converge at the global minumum. We formally state this in Theorem \ref{th:optimality} and the proof can be found in Appendix \ref{app:proof2}. We first define another property  similar to strict monotonicity. 

\parheader{Property 2.} $\forall t_1\ t_2\ t_3, \ (t_1<t_2) \text{ and } (t_3>0) \text{ implies } (t_1 \otimes t_3 < t_2 \otimes t_3$). 

\begin{theorem}
\label{th:optimality}
For any CLN model $M_F$ constructed from a formula, $F$, by the procedure shown
in the proof of Theorem 1, if $F$ is the conjunction of multiple linear equalities then any local minimum of $M_F$ is the global minimum, as long as the t-norm used in building $M_F$ satisfies Property 2.
\end{theorem}
\section{Loop Invariant Learning}
\label{sec:implementation}

\begin{figure}[t]
  \centering
  \includegraphics[width=0.9\linewidth]{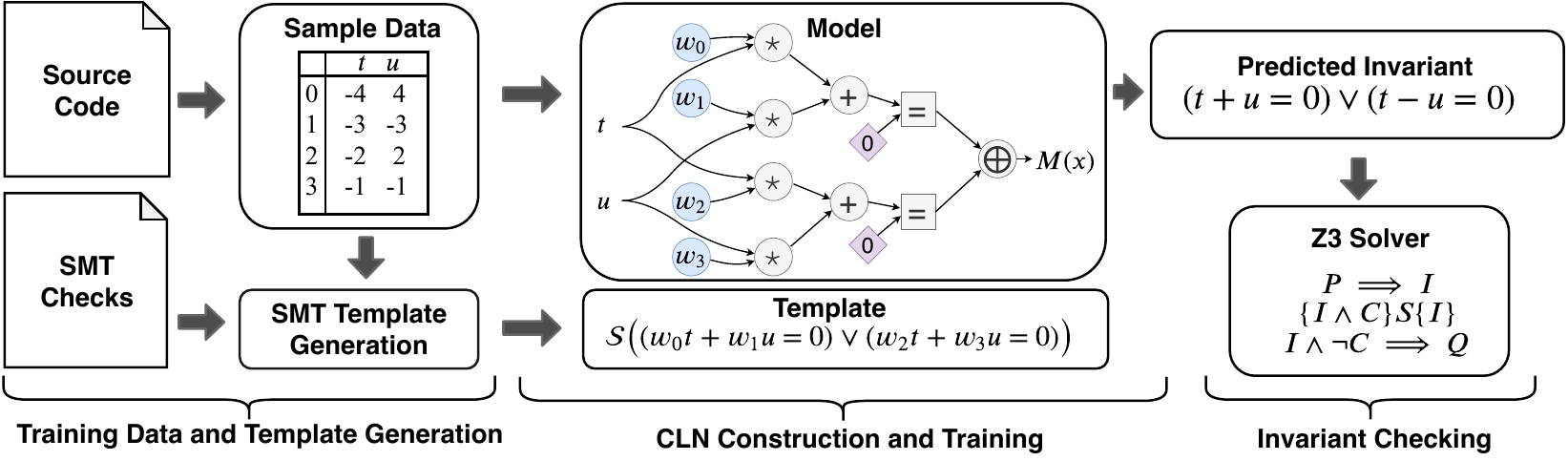}
  \setlength{\belowcaptionskip}{-10pt}
  \vspace{-0pt}
  \caption{\label{fig:arch}System architecture and CLN construction from SMT templates.}
  \vspace{-0pt}
\end{figure}

We use CLNs to implement a new inference system for loop invariants, CLN2INV, which learns invariants directly from execution traces. Figure \ref{fig:arch} provides an overview of the 
architecture.

\parheader{Training Data Generation.} 
We generate training data by running the program repeatedly on a set of randomly initialized inputs that satisfy the preconditions. Unconstrained variables are initialized from a uniform distribution, and variables with precondition constraints are initialized from a uniform distribution within their constraints. All program variables are recorded before each execution of the loop and after the loop terminates. 

\parheader{Template Generation.}
We encode the template using information gathered through static analysis. We collect useful information such as constants found in the program code along with the termination condition. Our analysis also strengthens the precondition and weakens the post-condition to constrain the problem as tightly as possible. For instance, unconstrained variables can be constrained to ensure the loop executes. In Appendix \ref{app:precondition_proof}, we prove this approach maintains soundness as solutions to the constrained problem can be used to reconstruct full solutions.

We generate bounds for individual variables (e.g., $t \geq b$) as well as multivariable polynomial constraints 
(e.g., $w_1 t_1 + w_2 t_2 = b$). Constants are optionally placed in constraints based on the static analysis and execution data (i.e. if a variable is initialized to a constant and never changes). We then compose template formulas from the collection of constraints by selecting a subset and joining them with conjunctions or disjunctions.

\parheader{CLN Construction and Training.} Once a template formula has been generated, a CLN is constructed from the template using the formulation in \S\ref{sec:methodology}.
As an optimization, we represent equality constraints as Gaussian-like functions that 
retain a global maximum when the constraint is satisfied as discussed in Appendix \ref{app:gaussian}. We then train the model using the collected execution traces.

\parheader{Invariant Checking.} Invariant checking is performed using SMT solvers such as Z3 \citep{de2008z3}. After the CLN for a formula template has been trained, the SMT formula for the loop invariant is recovered by normalizing the learned parameters. The invariant is checked against the pre, post, and recursion conditions as described in \S\ref{sec:back_invariants}.

\section{Experiments}
\label{sec:eval}

We compare the performance of CLN2INV with two existing methods and demonstrate the efficacy of the method on several more difficult problems. Finally, we conduct two ablation studies to justify our design choices.

\parheader{Test Environment.} All experiments are performed on an Ubuntu 18.04 server with an Intel Xeon E5-2623 v4 2.60GHz CPU, 256Gb of memory, and an Nvidia GTX 1080Ti GPU.

\parheader{System Configuration.} We implement CLNs in PyTorch and use the Adam optimizer for training with learning rate $0.01$ \citep{paszke2017automatic,kingma2014adam}. Because of the initialization dependency of neural networks, 
the CLN training randomly  restart if the model does not reach termination within $2,000$ epochs. Learnable parameters are initialized from a uniform distribution in the range [-1, 1], which we found works well in practice.  

\parheader{Test Dataset.} We use the same benchmark used in the evaluation of Code2Inv. We have removed nine invalid programs from Code2Inv's benchmark and test on the remaining 124. The removed programs are invalid because there are inputs which satisfy the precondition but result in a violation of the post-condition after the loop execution. 
The benchmark consists of loops expressed as C code and corresponding SMT files. Each loop can have nested if-then-else blocks (without nested loops). 
Programs in the benchmark may also have uninterpreted functions (emulating external function calls) in branches or loop termination conditions. 

\subsection{Comparison to existing solvers}

\begin{figure}[t]
  \centering
  \begin{subfigure}[b]{0.4\columnwidth}
    \includegraphics[width=\linewidth]{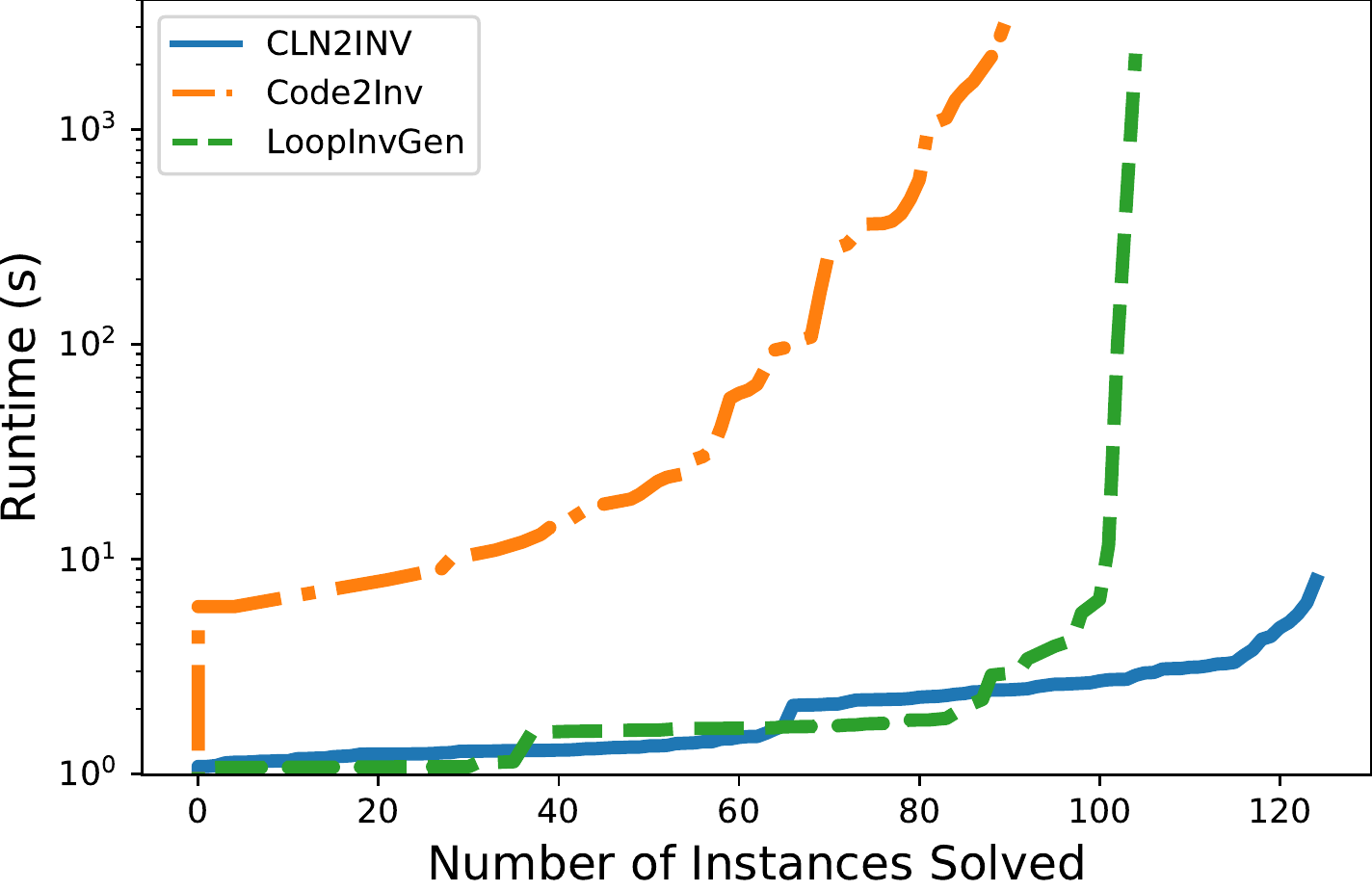}
    \caption{\label{fig:runtime}{\rm Runtime performance.}}
  \end{subfigure}
  \quad\quad\quad
  \begin{subfigure}[b]{0.4\columnwidth}
    \includegraphics[width=\linewidth]{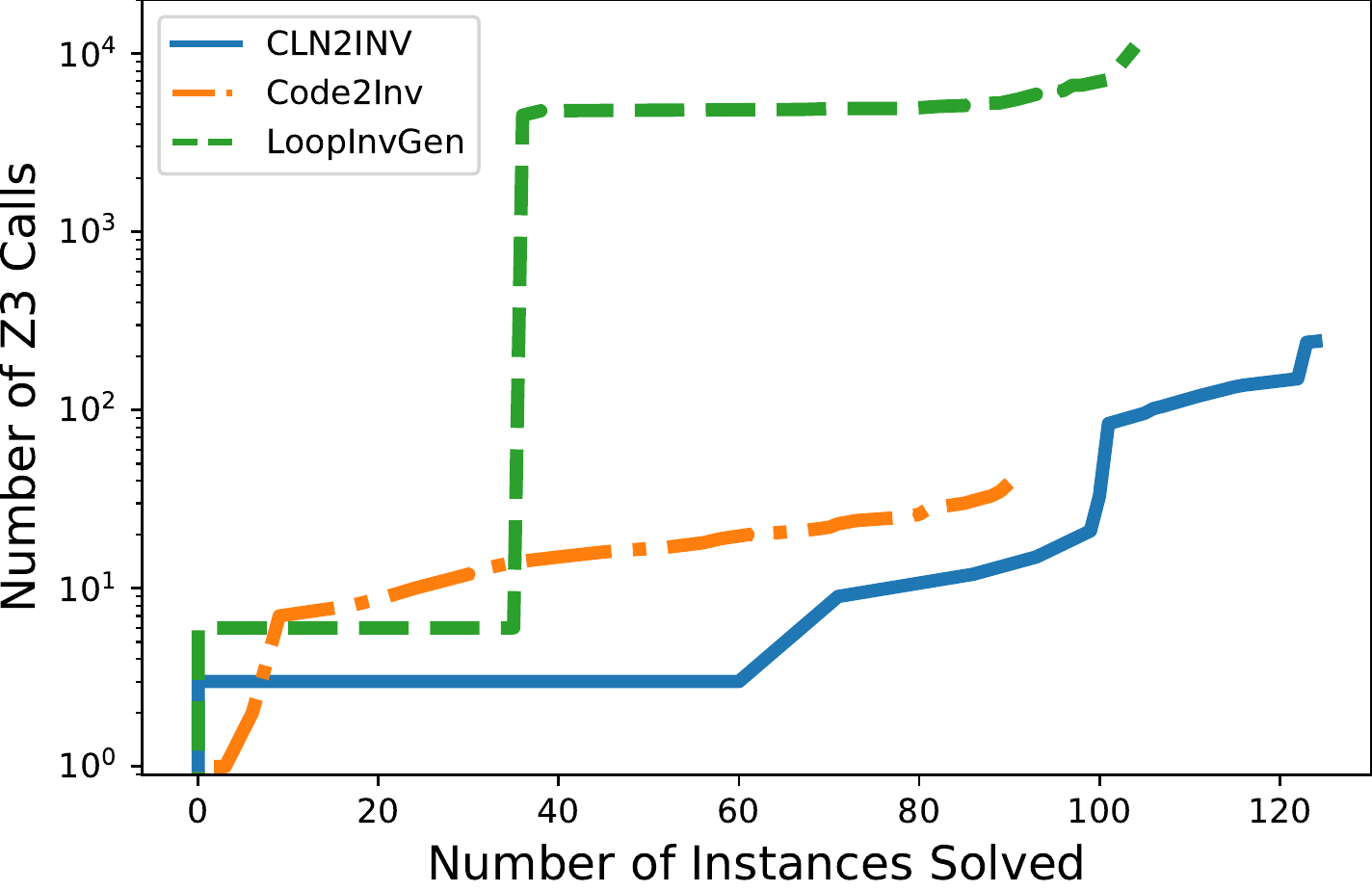}
    \caption{\label{fig:SMT}{\rm SMT solver calls.}}
  \end{subfigure}
 \setlength{\belowcaptionskip}{-10pt}
  \caption{\label{fig:working_ex}Performance evaluation.}
\end{figure}

\parheader{Performance Evaluation.} We compare CLN2INV to two state-of-the-art methods: Code2Inv (based on neural code representation and reinforcement learning) and LoopInvGen (PAC learning over synthesized CNF formulas)~\citep{si2018learning, padhi2016data}.  We limit each method to one hour per problem in the same format as the SyGuS Competition \citep{sygus2018}. Table \ref{tbl:summary} summarizes the results of the evaluation.  CLN2INV is able to solve all \textbf{124} problems in the benchmark. LoopInvGen solves \textbf{104} problems while Code2inv solves \textbf{90}.\footnote{The Code2Inv authors originally reported solving 92 problems with the same one hour timeout. We believe that the difference might be caused by changes in the testing environment or randomized model initialization.}

\parheader{Runtime Comparison.} Figure \ref{fig:runtime} shows the measured runtime on each evaluated system. CLN2INV solves problems in \textbf{1.1} second on average, which is over \textbf{40}$\times$ faster than LoopInvGen, the second fastest system in the evaluation. In general, CLN2INV has similar performance to LoopInvGen on simple problems, but is able to scale efficiently to  complex problems.

\parheader{Z3 Solver Utilization.} Figure \ref{fig:SMT} shows the number of Z3 calls made by each method. For almost all problems, CLN2INV requires fewer Z3 calls than the other systems, although for some difficult problems it uses more Z3 calls than Code2Inv. While CLN2INV makes roughly twice as many Z3 solver calls as Code2Inv on average, it is able to generate and test candidate loop invariants  over 250$\times$ faster on average.

\parheader{Performance Summary} Table \ref{tbl:summary} summarizes results of the performance evaluation. CLN2INV has the lowest time spent per problem making it the most practical approach. Code2Inv require more time on average per problem, but minimizes the number of calls made to an SMT solver. LoopInvGen is efficient at generating a large volume of guesses for the SMT solver. CLN2INV achieves a balance by producing quality guesses quickly allowing it to solve problems efficiently.

\begin{table}[h]
\caption{ \label{tbl:summary}Results and summary statistics for performance evaluation. }
\centering
\begin{tabular}{l*{4}{r}}
\toprule
\textbf{Method} &
{\specialcellbold{Number Solved}} &
{\specialcellbold{Avg Time (s)}} &
{\specialcellbold{Avg Z3 Calls}} & 
{\specialcellbold{Time/Z3 Call (s)}}\\
\midrule
Code2Inv & 90 & 266.71 & 16.62 & 50.89 \\
 LoopInvGen & 104 & 45.11 & 3,605.43 & 0.08\\
 CLN2INV& 124 & 1.07 & 31.77 & 0.17 \\
\bottomrule
\end{tabular}

\end{table}

\subsection{More Difficult Loop Invariants}
\label{sec:challenge}
We construct \NComplexInvariants{} additional problems
to demonstrate CLN2INV's ability to infer  complex loop invariants. We design these problems to have two key characteristics, which are absent in the Code2Inv dataset: (i) they require invariants involving conjunctions and disjunctions of multivariable constraints, and (ii) the invariant cannot easily be identified by inspecting the precondition, termination condition, or post-condition. CLN2INV is able to find correct invariants for all \NComplexInvariants{} problems in less than 20 seconds, while Code2Inv and LoopInvGen time out after an hour (see Appendix \ref{app:ex_loop_challenge}).

\subsection{Ablation Studies}

\begin{figure}[H]
\centering
\begin{subfigure}[b]{0.4\textwidth}
\includegraphics[width=\linewidth]{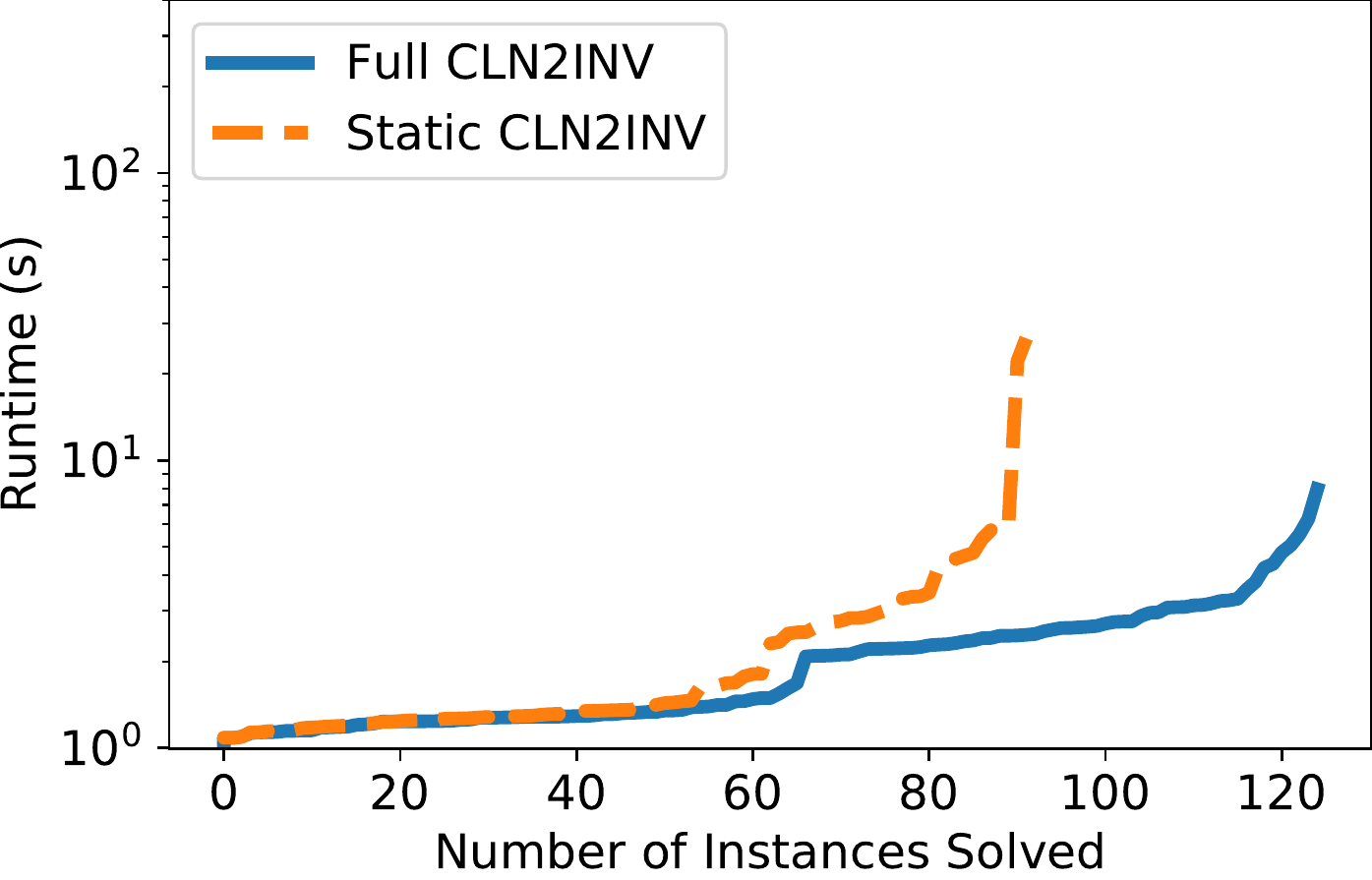}
\caption{\label{fig:abl_time}Runtime performance. }
\end{subfigure}
\quad\quad\quad
\begin{subfigure}[b]{0.4\textwidth}
\includegraphics[width=\linewidth]{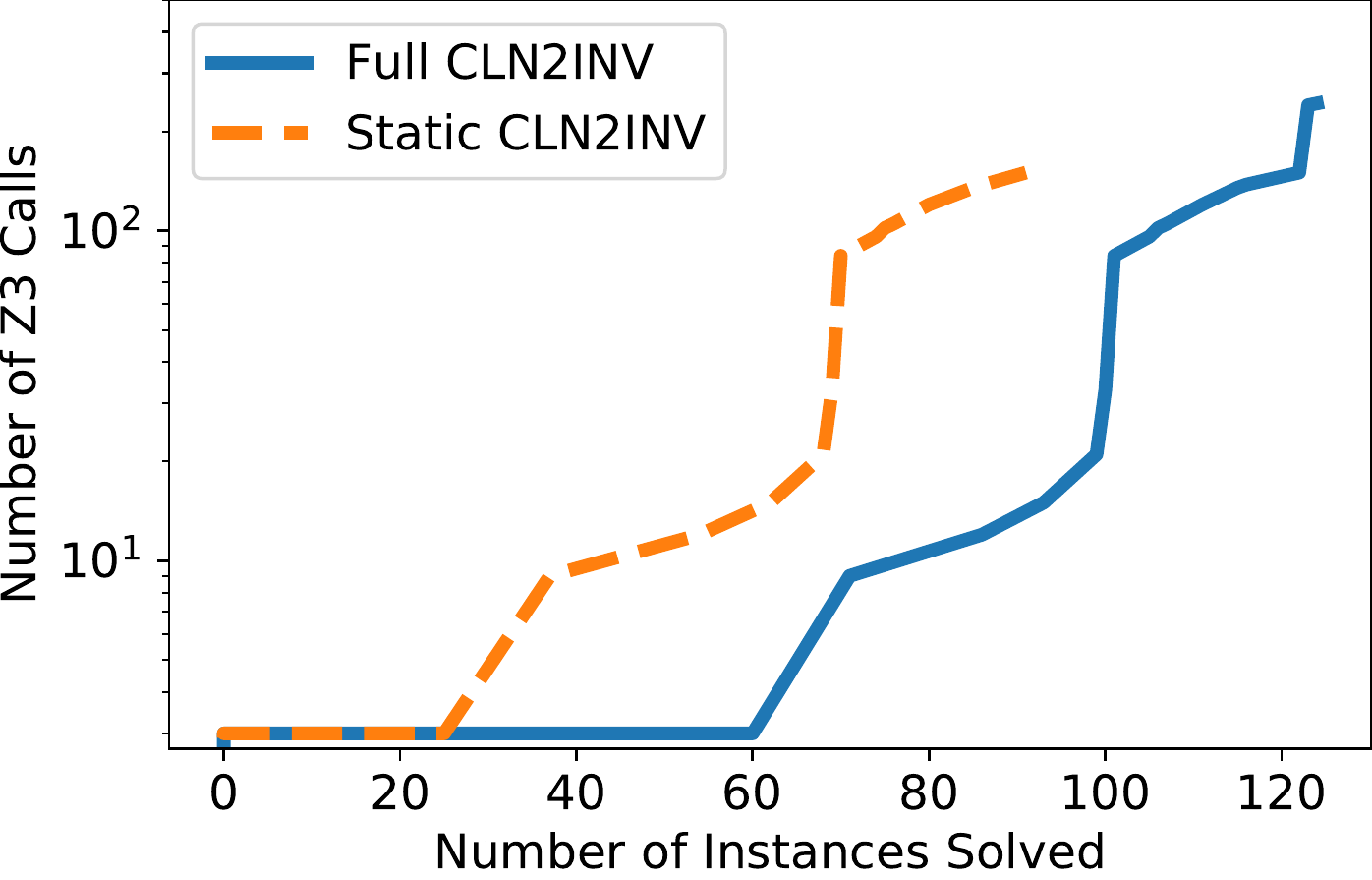}
\caption{\label{fig:abl_z3}SMT solver calls.}
\end{subfigure}
 \setlength{\belowcaptionskip}{-10pt}
\caption{\label{fig:abl}Ablation study: Comparing Static vs Trained Models}
\end{figure}

\parheader{Effect of CLN Training on Performance.}
CLN2INV relies on a combination of heuristics using static analysis and learning formulas from execution traces to  correctly infer loop invariants. In this ablation we disable model training and limit CLN2INV to static models with no learnable parameters. The static CLN2INV solves 91 problems in the dataset. Figure \ref{fig:abl} shows a comparison of full CLN2INV with one limited to static models. CLN2INV's performance with training disabled shows that a large number of problems in the dataset are relatively simple and invariants can be inferred from basic heuristics. However, for more difficult problems, the ability of CLNs to learn SMT formulas is key to successfully finding correct invariants.

\parheader{Comparing t-norms.}
We compare the effect of different t-norms (Godel, Lukasiewicz, and Product) on convergence time in Table \ref{tbl:tnorms}. All t-norms have very similar performance when used for conjunctions, but the product t-conorm converges faster than the other t-conorms on average. 
See Appendix \ref{app:norms} for more details.

 \begin{table}[H]
 \centering
 \small
 \caption{ \label{tbl:tnorms}Table with average iterations to convergence (average taken over 5 runs) }
 \begin{tabular}{l*{3}{r}}
 \toprule
 \textbf{Problem} &
 {\specialcellbold{Godel (iterations)}} &
 {\specialcellbold{Lukasiewicz (iterations)}} &
 {\specialcellbold{Product (iterations)}}\\
 \midrule
 Conjunction & 967 & 966 & 966  \\
  Disjunction & 1,074 & 1,221 & 984 \\
 \bottomrule
 \end{tabular}
 \end{table}
 \vspace{-20pt}

\section{conclusion}
\label{sec:conclusion}

We develop a novel neural architecture that explicitly and precisely learns SMT formulas by construction. We achieve this by introducing a new sound and complete semantic mapping for SMT that enables learning formulas through backpropagation. 
We use CLNs to implement a loop invariant inference system, CLN2INV, that is the first to solve all theoretically solvable problems in the Code2Inv benchmark %
and takes only 1.1 seconds on average. We believe that the CLN architecture will also be beneficial for other domains that require learning SMT formulas.

\bibliography{main}
\bibliographystyle{iclr2020_conference}

\begin{appendices}
\newpage

\section{Proof of limit of $\C(t=u)$}
\label{app:limit-equality}
\[
\lim_{\substack{\epsilon \rightarrow 0^+ \\ B \cdot \epsilon \rightarrow +\infty}}\!\!\!\! \C(t = u) 
\;\; =
\lim_{\substack{\epsilon \rightarrow 0^+ \\ B \cdot \epsilon \rightarrow +\infty}}\!\! \frac{1}{1+e^{-B(t-u+\epsilon)}} \otimes \frac{1}{1+e^{B(t-u-\epsilon)}} = \left\{\begin{array}{lr}
1 & t = u \\
0 & t \neq u
\end{array}\right.\]
\begin{proof}
Let $f(t, u;B,\epsilon) =  \frac{1}{1+e^{-B(t-u+\epsilon)}}$
and $g(t, u;B,\epsilon) = \frac{1}{1+e^{B(t-u-\epsilon)}}$. Then what we want to prove becomes 
\[\lim_{\substack{\epsilon \rightarrow 0^+ \\ B \cdot \epsilon \rightarrow +\infty}} \!\!\!\! 
\left( f(t,u;B,\epsilon) \ \otimes\ g(t,u;B,\epsilon) \right) = \left\{
\begin{array}{cc}
    1 & t = u \\
    0 & t \neq u
\end{array}\right\}\]
Because all $f,g,\otimes$ are continuous in their domain, we have
\[\lim_{\substack{\epsilon \rightarrow 0^+ \\ B \cdot \epsilon \rightarrow +\infty}}\!\!\!\! \left( f(t,u;B,\epsilon) \ \otimes\ g(t,u;B,\epsilon) \right) = \left(\lim_{\substack{\epsilon \rightarrow 0^+ \\ B \cdot \epsilon \rightarrow +\infty}}\!\!\!\!f(t,u;B,\epsilon)\right)
\otimes
\left(\lim_{\substack{\epsilon \rightarrow 0^+ \\ B \cdot \epsilon \rightarrow +\infty}}\!\!\!\! g(t,u;B,\epsilon)\right)\]
Using basic algebra, we get
\begin{align*}
\lim_{\substack{\epsilon \rightarrow 0^+ \\ B \cdot \epsilon \rightarrow +\infty}} f(t,u;B,\epsilon) = \left\{
\begin{array}{cc}
    1 & t \geq u \\
    0 & t < u
\end{array}\right. && 
\lim_{\substack{\epsilon \rightarrow 0^+ \\ B \cdot \epsilon \rightarrow +\infty}} g(t,u;B,\epsilon) = \left\{
\begin{array}{cc}
    1 & t \leq u \\
    0 & t > u
\end{array}\right.
\end{align*}
Combing these results, we have
\[\left(\lim_{\substack{\epsilon \rightarrow 0^+ \\ B \cdot \epsilon \rightarrow +\infty}}\!\!\!\!f(t,u;B,\epsilon)\right)
\otimes
\left(\lim_{\substack{\epsilon \rightarrow 0^+ \\ B \cdot \epsilon \rightarrow +\infty}}\!\!\!\! g(t,u;B,\epsilon)\right) = \left\{
\begin{array}{cc}
    0 \otimes 1 & t<u \\
    1 \otimes 1 & t=u \\
    1 \otimes 0 & t>u
\end{array}\right.\]
For any t-norm, we have $0 \otimes 1 = 0$, $1 \otimes 1 = 1$, and $1 \otimes 0 = 0$. Put it altogether, we have
\[\lim_{\substack{\epsilon \rightarrow 0^+ \\ B \cdot \epsilon \rightarrow +\infty}} \!\!\!\! 
\left( f(t,u;B,\epsilon) \ \otimes\ g(t,u;B,\epsilon) \right) = \left\{
\begin{array}{cc}
    1 & t = u \\
    0 & t \neq u
\end{array}\right.\]
which concludes the proof.
\end{proof}

\section{Proof of Theorem 1}
\label{app:proof1}
\parheader{Theorem 1.}
For any quantifier-free linear SMT formula $F$, there exists CLN model $M$, such that
\begin{align}
    \forall \bm{x}\ B\ \epsilon, \ 0 \leq M(\bm{x};B,\epsilon) \leq 1
\end{align}
\begin{align}
    \forall \bm{x}, \ F(\bm{x})=True \iff \lim_{\substack{\epsilon \rightarrow 0^+ \\ B\cdot\epsilon\rightarrow\infty}} M(\bm{x};B,\epsilon)=1
\end{align}
\begin{align}
    \forall \bm{x}, \ F(\bm{x})=False \iff \lim_{\substack{\epsilon \rightarrow 0^+ \\ B\cdot\epsilon\rightarrow\infty}} M(\bm{x};B,\epsilon)=0
\end{align}
as long as the t-norm used in building $M$ satisfies Property 1.

\begin{proof}
For convenience of the proof, we first remove all $<$, $\leq$, $=$ and $\neq$ in $F$, by transforming $t<u$ into $\neg (t \geq u)$, $t \leq u$ into $\neg (t>u)$, $t=u$ into $(t \geq u) \land \neg (t>u)$, and $t \neq u$ into $(t>u) \lor \neg (t \geq u)$. Now the only operators that $F$ may contain are $>, \geq, \land, \lor, \neg$. We prove Theorem 1 by induction on the constructor of formula $F$. In the following proof, we construct model $M$ given $F$ and show it satisfied Eq.(1)(2). We leave the proof for why $M$ also satisfied Eq.(3) to readers.

\parheader{Atomic Case.}
When $F$ is an atomic clause, then $F$ will be in the form of $\bm{x}*W + b > 0$ or $\bm{x}*W + b \geq 0$. For the first case, we construct a linear layer with weight $W$ and bias $b$ followed by a sigmoid function scaled with factor $B$ and right-shifted with distance $\epsilon$. For the second case, we construct the same linear layer followed by a sigmoid function scaled with factor $B$ and left-shifted with distance $\epsilon$. Simply evaluating the limits for each we arrive at $$\forall \bm{x}, F(\bm{x})=True \iff \lim_{\substack{\epsilon \rightarrow 0^+\\B\cdot\epsilon\rightarrow\infty}} M(\bm{x};B,\epsilon)=1 $$ And from the definition of sigmoid function we know $0 \leq M(\bm{x};B,\epsilon) \leq 1$. 

\parheader{Negation Case.} If $F=\neg F'$, from the induction hypothesis, $F'$ can be represented by models $M'$ satisfying Eq.(1)(2)(3). Let $p'$ output node of $M'$. We add a final output node $p = 1 - p'$. So $M(\bm{x};B,\epsilon) = 1 - M'(\bm{x};B,\epsilon)$. Using the induction hypothesis $0 \leq M'(\bm{x};B,\epsilon) \leq 1$, we conclude Eq.(1) $0 \leq M(\bm{x};B,\epsilon) \leq 1$.

Now we prove the ``$\implies$'' side of Eq.(2). If $F(\bm{x})=True$, then $F'(\bm{x})=False$. From the induction hypothesis, we know $lim_{\substack{\epsilon \rightarrow 0^+\\B\cdot\epsilon\rightarrow\infty}} M'(\bm{x};B,\epsilon) = 0$. So $$\lim_{\substack{\epsilon \rightarrow 0^+\\B\cdot\epsilon\rightarrow\infty}} M(\bm{x};B,\epsilon) = \lim_{\substack{\epsilon \rightarrow 0^+\\B\cdot\epsilon\rightarrow\infty}} 1 - M'(\bm{x};B,\epsilon)= 1 - 0 = 1$$

Next we prove the ``$\impliedby$'' side. If $\lim_{\substack{\epsilon \rightarrow 0^+\\B\cdot\epsilon\rightarrow\infty}} M(\bm{x};B,\epsilon) = 1$, we have 
\[\lim_{\substack{\epsilon \rightarrow 0^+\\B\cdot\epsilon\rightarrow\infty}} M'(\bm{x};B,\epsilon) = \lim_{\substack{\epsilon \rightarrow 0^+\\B\cdot\epsilon\rightarrow\infty}} 1 - M(\bm{x};B,\epsilon)= 1 - 1 = 0\]
From the induction hypothesis we know that $F'(\bm{x})=False$. So $F(\bm{x})=\neg F'(\bm{x})=True$.

\parheader{Conjunction Case.} If $F=F_1 \land F_2$, from the induction hypothesis, $F_1$ and $F_2$ can be represented by models $M_1$ and $M_2$, such that both $(F_1,M_1)$ and $(F_2,M_2)$ satisfy Eq.(1)(2)(3). Let $p_1$ and $p_2$ be the output nodes of $M_1$ and $M_2$. We add a final output node $p = p_1 \otimes p_2$. So $M(\bm{x};B,\epsilon)=M_1(\bm{x};B,\epsilon) \otimes M_2(\bm{x};B,\epsilon)$. Since $(\otimes)$ is continuous and so are $M_1(\bm{x};B,\epsilon)$ and $M_2(\bm{x};B,\epsilon)$, we know their composition $M(\bm{x};B,\epsilon)$ is also continuous. (Readers may wonder why $M1(\bm{x};B,\epsilon)$ is continuous. Actually the continuity of $M(\bm{x};B,\epsilon)$ should be proved inductively like this proof itself, and we omit it for brevity.) From the definition of ($\otimes)$, we have Eq.(1) $0 \leq M(\bm{x};B,\epsilon) \leq 1$.

Now we prove the $\implies$ side of Eq.(2). For any $\bm{x}$, if $F(\bm{x})=True$ which means both $F_1(\bm{x})=True$ and $F_2(\bm{x})=True$, from the induction hypothesis we know that $\lim_{\substack{\epsilon \rightarrow 0^+\\B\cdot\epsilon\rightarrow\infty}} M_1(\bm{x};B,\epsilon)=1$ and $\lim_{\substack{\epsilon \rightarrow 0^+\\B\cdot\epsilon\rightarrow\infty}} M_2(\bm{x};B,\epsilon)=1$. Then $$\lim_{\substack{\epsilon \rightarrow 0^+\\B\cdot\epsilon\rightarrow\infty}} M(\bm{x};B,\epsilon)=\lim_{\substack{\epsilon \rightarrow 0^+\\B\cdot\epsilon\rightarrow\infty}}M_1(\bm{x};B,\epsilon) \otimes M_2(\bm{x};B,\epsilon) = 1 \otimes 1 = 1$$ 

Then we prove the $\impliedby$ side. From the induction hypothesis we know that $M_1(\bm{x};B,\epsilon) \leq 1$ and $M_2(\bm{x};B,\epsilon) \leq 1$. From the non-decreasing property of t-norms (see \S\ref{sec:bl_logic}), we have $$M_1(\bm{x};B,\epsilon) \otimes M_2(\bm{x};B,\epsilon)\;\; \leq\;\; M_1(\bm{x};B,\epsilon) \otimes 1$$ 

Then from the consistency property and the commutative property, we have 
$$M_1(\bm{x};B,\epsilon) \otimes 1 = M_1(\bm{x};B,\epsilon)$$
Put them altogether we get
\[M(\bm{x};B,\epsilon) \;\; \leq \;\; M_1(\bm{x};B,\epsilon) \;\; \leq \;\; 1\]
Because we know $\lim_{\substack{\epsilon \rightarrow 0^+\\B\cdot\epsilon\rightarrow\infty}} M(\bm{x};B,\epsilon)=1$, according to the squeeze theorem in calculus, we get $$\lim_{\substack{\epsilon \rightarrow 0^+\\B\cdot\epsilon\rightarrow\infty}} M_1(\bm{x};B,\epsilon)=1$$ From the induction hypothesis, we know that $F_1(\bm{x})=True$. We can prove $F_2(\bm{x})=True$ in the same manner. Finally we have $F(\bm{x})=F_1(\bm{x}) \land F_2(\bm{x})=True$.

\parheader{Disjunction Case.} For the case $F=F_1 \lor F_2$, we construct $M$ from $M_1$ and $M_2$ as we did in the conjunctive case. This time we let the final output node be $p = p_1 \oplus p_2$. From the continuity of ($\otimes$) and the definition of ($\oplus$) ($t \oplus u = 1 - (1-t) \otimes (1-u)$), $(\oplus)$ is also continuous. We conclude $M(\bm{x};B,\epsilon)$ is also continuous and $0 \leq M(\bm{x};B,\epsilon) \leq 1$ by the same argument as $F=F_1 \land F_2$. 

Now we prove the ``$\implies$'' side of Eq.(2). For any assignment $\bm{x}$, if $F(\bm{x})=True$ which means $F_1(\bm{x})=True$ or $F_2(\bm{x})=True$. Without loss of generality, we assume $F_1(\bm{x})=True$. From the induction hypothesis, we know $\lim_{\substack{\epsilon \rightarrow 0^+\\B\cdot\epsilon\rightarrow\infty}} M_1(\bm{x};B,\epsilon)=1$. 

For any ($\oplus$) and any $0 \leq t,t' \leq 1$, if $t \leq t'$, then 
$$t \oplus u = 1 - (1-t) \otimes (1-u) \;\; \leq \;\; 1 - (1-t') \otimes (1-u) = t' \oplus u$$ 

Using this property and the induction hypothesis $M_2(\bm{x};B,\epsilon) \geq 0$, we have 
$$M_1(\bm{x};B,\epsilon) \oplus 0 \;\; \leq \;\; M_1(\bm{x};B,\epsilon) \oplus M_2(\bm{x};B,\epsilon) = M(\bm{x};B,\epsilon)$$ 

From the induction hypothesis we also have $M_1(\bm{x};B,\epsilon) \leq 1$. Using the definition of ($\oplus$) and the consistency of ($\otimes$) ($0 \otimes x = 0$), we get $M_1(\bm{x};B,\epsilon) \oplus 0 = M_1(\bm{x};B,\epsilon)$. Put them altogether we get
\[M_1(\bm{x};B,\epsilon) \;\; \leq \;\; M(\bm{x};B,\epsilon) \leq 1\]
Because we know $\lim_{\substack{\epsilon \rightarrow 0^+\\B\cdot\epsilon\rightarrow\infty}} M_1(\bm{x};B,\epsilon)=1$, according to the squeeze theorem in calculus, we get $\lim_{\substack{\epsilon \rightarrow 0^+\\B\cdot\epsilon\rightarrow\infty}} M(\bm{x};B,\epsilon)=1$.

Then we prove the  ``$\impliedby$'' side. Here we need to use the existence of limit: $$\lim_{\substack{\epsilon \rightarrow 0^+\\B\cdot\epsilon\rightarrow\infty}} M(\bm{x};B,\epsilon)$$
This property can be proved by induction like this proof itself, thus omitted for brevity. 

Let $$c_1 = \lim_{\substack{\epsilon \rightarrow 0^+\\B\cdot\epsilon\rightarrow\infty}} M_1(\bm{x};B,\epsilon)$$ and  $$c_2 = \lim_{\substack{\epsilon \rightarrow 0^+\\B\cdot\epsilon\rightarrow\infty}} M_2(\bm{x};B,\epsilon)$$
Then $$\lim_{\substack{\epsilon \rightarrow 0^+\\B\cdot\epsilon\rightarrow\infty}} M(\bm{x};B,\epsilon)=\lim_{\substack{\epsilon \rightarrow 0^+\\B\cdot\epsilon\rightarrow\infty}}M_1(\bm{x};B,\epsilon) \oplus M_2(\bm{x};B,\epsilon) = c_1 \oplus c_2 = 1 - (1-c_1) \otimes (1-c_2)$$
Since we have $\lim_{\substack{\epsilon \rightarrow 0^+\\B\cdot\epsilon\rightarrow\infty}} M(\bm{x};B,\epsilon) = 1$, we get $$(1-c_1) \otimes (1-c_2)=0$$
Using Property 1 of ($\otimes$) (defined in \S\ref{sec:methodology}), we have $c_1=1 \lor c_2=1$. Without loss of generality, we assume $c_1=1$. From the induction hypothesis, we know that $F_1(\bm{x})=True$. Finally, $F(\bm{x})=F_1(\bm{x}) \lor F_2(\bm{x})=True$.

\end{proof}

Careful readers may have found that if we use the continuous mapping function $\C$ in \S\ref{sec:logic}, then we have another perspective of the proof above, which can be viewed as two interwoven parts. The first part is that we proved the following lemma.
\begin{corollary}
For any quantifier-free linear SMT formula $F$,
\[\forall \bm{x}\ B \ \epsilon, \ 0 \leq \C(F;B,\epsilon)(\bm{x}) \leq 1\]
\[\forall \bm{x}, \ F(\bm{x})=True \iff \lim_{\substack{\epsilon \rightarrow 0^+\\B\cdot\epsilon\rightarrow\infty}} \C(F;B,\epsilon)(\bm{x})=1\]
\[\forall \bm{x}, \ F(\bm{x})=False \iff \lim_{\substack{\epsilon \rightarrow 0^+\\B\cdot\epsilon\rightarrow\infty}} \C(F;B,\epsilon)(\bm{x})=0\]
\end{corollary}
Corollary 1 indicates the soundness of $\C$. The second part is that we construct a CLN model given $\C(F)$. In other words, we translate $\C(F)$ into vertices in a computational graph composed of differentiable operations on continuous truth values.

\newpage
\section{Proof of Theorem 2}
\label{app:proof2}

\parheader{Theorem 2.} For any CLN model $M_F$ constructed from a formula, $F$, by the procedure shown
in the proof of Theorem 1, if $F$ is the conjunction of multiple linear equalities then any local minima of $M_F$ is the global minima, as long as the t-norm used in building $M_F$ satisfies Property 2.

\begin{proof}
Since $F$ is the conjunction of linear equalities, it has the form
\[F = \bigwedge_{i=1}^n (\sum_{j=1}^{l_i} w_{ij}t_{ij} = 0)\]
Here $\mathbf{W}=\{w_{ij}\}$ are the learnable weights, and $\{t_{ij}\}$ are terms (variables). We omit the bias $b_i$ in the linear equalities, as the bias can always be transformed into a weight by adding a constant of 1 as a term. For convenience, we define $f(x)=\C(x=0)= \frac{1}{1+e^{-B(x+\epsilon)}} \otimes \frac{1}{1+e^{B(x-\epsilon)}}$.

Given an assignment $\bm{x}$ of the terms $\{t_{ij}\}$, if we construct our CLN model $M_F$ following the procedure shown in the proof of Theorem 1, the output of the model will be
\[M(\bm{x};\mathbf{W},B,\epsilon) = \bigotimes_{i=1}^n f(\sum_{j=1}^{l_i} w_{ij} t_{ij})\]
When we train our CLN model, we have a collection of $m$ data points $\{t_{ij1}\},\{t_{ij2}\},...,\{t_{ijm}\},$ which satisfy formula $F$. If $B$ and $\epsilon$ are fixed (unlearnable), then the loss function will be
\begin{align}
L(\mathbf{W}) = \sum_{k=1}^m \mathcal{L}(M(\bm{x};\mathbf{W},B,\epsilon)) = \sum_{k=1}^m \mathcal{L}(\bigotimes_{i=1}^n f(\sum_{j=1}^{l_i} w_{ij} t_{ijk}))
\end{align}
Suppose $\mathbf{W}^*=\{w_{ij}^*\}$ is a local minima of $L(\mathbf{W})$. We need to prove $\mathbf{W}^*$ is also the global minima. To prove this, we use the definition of a local minima. That is, 
\begin{align}
\exists \delta>0, \ \forall \mathbf{W}, \ ||\mathbf{W}-\mathbf{W}^*|| \leq \delta \implies L(\mathbf{W}) \geq L(\mathbf{W}^*)
\end{align}
For convenience, we denote $u_{ik} = \sum_{j=1}^{l_i} w_{ij} t_{ijk}$. Then we rewrite Eq.(4) as 
\[L(\mathbf{W}) = \sum_{k=1}^m \mathcal{L}(\bigotimes_{i=1}^n f(u_{ik}))\]
If we can prove at $\mathbf{W}^*$, $\forall \  1 \leq i \leq n, \ 1 \leq k \leq m, \ u_{ik}=0$. Then because (\romannumeral 1) $f$ reaches its global maximum at $0$, (\romannumeral 2) the t-norm $(\otimes)$ is monotonically increasing, (\romannumeral 3) $\mathcal{L}$ is monotonically decreasing, we can conclude that $\mathbf{W}^*$ is the global minima.

Now we prove $\forall \  1 \leq i \leq n, \ 1 \leq k \leq m, \ u_{ik}=0$. Here we just show the case $i=1$. The proof for $i>1$ can be directly derived using the associativity of $(\otimes)$.

Let $\alpha_k = \bigotimes_{i=2}^n f(u_{ik})$. Since $f(x)>0$ for all $x \in R$, using Property 2 of our t-norm $(\otimes)$, we know that $\alpha_k>0$. Now the loss function becomes
\[L(\mathbf{W}) = \sum_{k=1}^m \mathcal{L}(f(u_{1k}) \otimes \alpha_k)\]
From Eq.(5), we have
\begin{align}
\exists 0<\delta'<1, \  \forall \gamma, \ |\gamma| \leq \delta' \implies \sum_{k=1}^m \mathcal{L}(f(u_{ik}(1+\gamma)) \otimes \alpha_k) \geq \sum_{k=1}^m \mathcal{L}(f(u_{ik}) \otimes \alpha_k)
\end{align}

Because (\romannumeral 1) $f(x)$ is an even function decreasing on $x>0$ (which can be easily proved), (\romannumeral 2) $(\otimes)$ is monotonically increasing, (\romannumeral 3) $\mathcal{L}$ is monotonically decreasing, for $-\delta' < \gamma < 0$, we have
\begin{align}
\sum_{k=1}^m \mathcal{L}(f(u_{ik}(1+\gamma)) \otimes \alpha_k) & = \sum_{k=1}^m \mathcal{L}(f(|u_{ik}(1+\gamma)|) \otimes \alpha_k) \leq \nonumber
\\ \sum_{k=1}^m \mathcal{L}(f(|u_{ik}|) \otimes \alpha_k) & = \sum_{k=1}^m \mathcal{L}(f(u_{ik}) \otimes \alpha_k)
\end{align}
Combing Eq.(6) and Eq.(7), we have 
\[\sum_{k=1}^m \mathcal{L}(f(u_{ik}(1+\gamma)) \otimes \alpha_k) = \sum_{k=1}^m \mathcal{L}(f(u_{ik}) \otimes \alpha_k)\]
Now we look back on Eq.(7). Since (\romannumeral 1) $\mathcal{L}$ is strictly decreasing, (\romannumeral 2) the t-norm we used here has Property 2 (see \S\ref{sec:methodology} for definition), (\romannumeral 3) $\alpha_k > 0$, the only case when ($=$) holds is that for all $1 \leq k \leq m$, we have $f(|u_{ik}(1+\gamma)|) = f(|u_{ik}|)$. Since $f(x)$ is strictly decreasing for $x \geq 0$, we have $|u_{ik}(1+\gamma)| = |u_{ik}|$. Finally because $-1 < -\delta' < \gamma < 0$, we have $u_{ik} = 0$.
\end{proof}

\section{Theorem 3 and the proof}
\label{app:precondition_proof}

\parheader{Theorem 3.} Given a program $C$: assume($P$); while ($LC$) \{$C$\} assert($Q$); \newline
If we can find a loop invariant $I'$ for program $C'$: assume($P\land LC$); while ($LC$) \{$C$\} assert($Q$); \newline
and $P\land \neg LC \implies Q$, then $I' \lor (P \land \neg LC)$ is a correct loop invariant for program $C$.

\begin{proof}
Since $I'$ is a loop invariant of $C'$, we have
\begin{align*}
    (P \land LC) \land LC \implies I'\ \ \ \ (a) && \{I' \land LC\}C\{I'\}\ \ \ \ (b) && I' \land \neg LC \implies Q\ \ \ \ (c)
\end{align*}
We want to prove $I' \lor (P \land \neg LC)$ is a valid loop invariant of C, which means
\begin{align*}
    P \land LC \implies I' \lor (P \land \neg LC) && \{(I' \lor (P \land \neg LC)) \land LC\}C\{I' \lor (P \land \neg LC)\} \\
    (I' \lor (P \land \neg LC)) \land \neg LC \implies Q
\end{align*}
We prove the three propositions separately. To prove $P \land LC \implies I' \lor (P \land \neg LC)$, we transform it into a stronger proposition $P \land LC \implies I'$, which directly comes from (a).

For $\{(I' \lor (P \land \neg LC)) \land LC\}C\{I' \lor (P \land \neg LC)\}$, after simplification it becomes $\{I' \land LC\}C\{I' \lor (P \land \neg LC)\}$, which is a direct corollary of (b).

For $(I' \lor (P \land \neg LC)) \land \neg LC \implies Q$, after simplification it will become two separate propositions, $I' \land \neg LC \implies Q$ and $P \land \neg LC \implies Q$. The former is exactly (c), and the latter is a known condition in the theorem.
\end{proof}

\section{Properties of Gaussian Function}
\label{app:gaussian}

We use a Gaussian-like function $S(t=u)=\exp(-\frac{(t-u)^2}{2\sigma_2})$ to represent equalities in our experiments. It has the following two properties. First, it preserves the original semantic of $=$ when $\sigma \rightarrow 0$, similar to the mapping $S(t=u)= \frac{1}{1+e^{-B(t-u+\epsilon)}} \otimes \frac{1}{1+e^{B(t-u-\epsilon)}}$ we defined in \S\ref{sec:logic}.
\[\lim_{\sigma \rightarrow 0^+} \exp(-\frac{(t-u)^2}{2\sigma^2}) = \left\{\begin{array}{cc}
    1 & t=u \\
    0 & t \neq u
\end{array}\right.\]
Second, if we view $S(t=u)$ as a function over $t-u$, then it reaches its only local maximum at $t-u=0$, which means the equality is satisfied.

\newpage
\section{More Difficult Loop Invariant}
\label{app:ex_loop_challenge}

\begin{figure}[h]
\vspace{-10pt}
\centering
\begin{subfigure}[b]{0.45\textwidth}
\centering
\begin{lstlisting}[style=myC,numbers=none]
//pre: t=-20/\u=-20
while (u != 0) {
  t++;
  if (u > 0)
    t = -u + 1;
  else 
    t = -u - 1;
}
//post: t=0
\end{lstlisting}
\vspace{5pt}
\caption{\label{fig:P11} Example Pseudocode}
\end{subfigure}
\qquad
\begin{subfigure}[b]{0.4\textwidth}
\includegraphics[width=\linewidth]{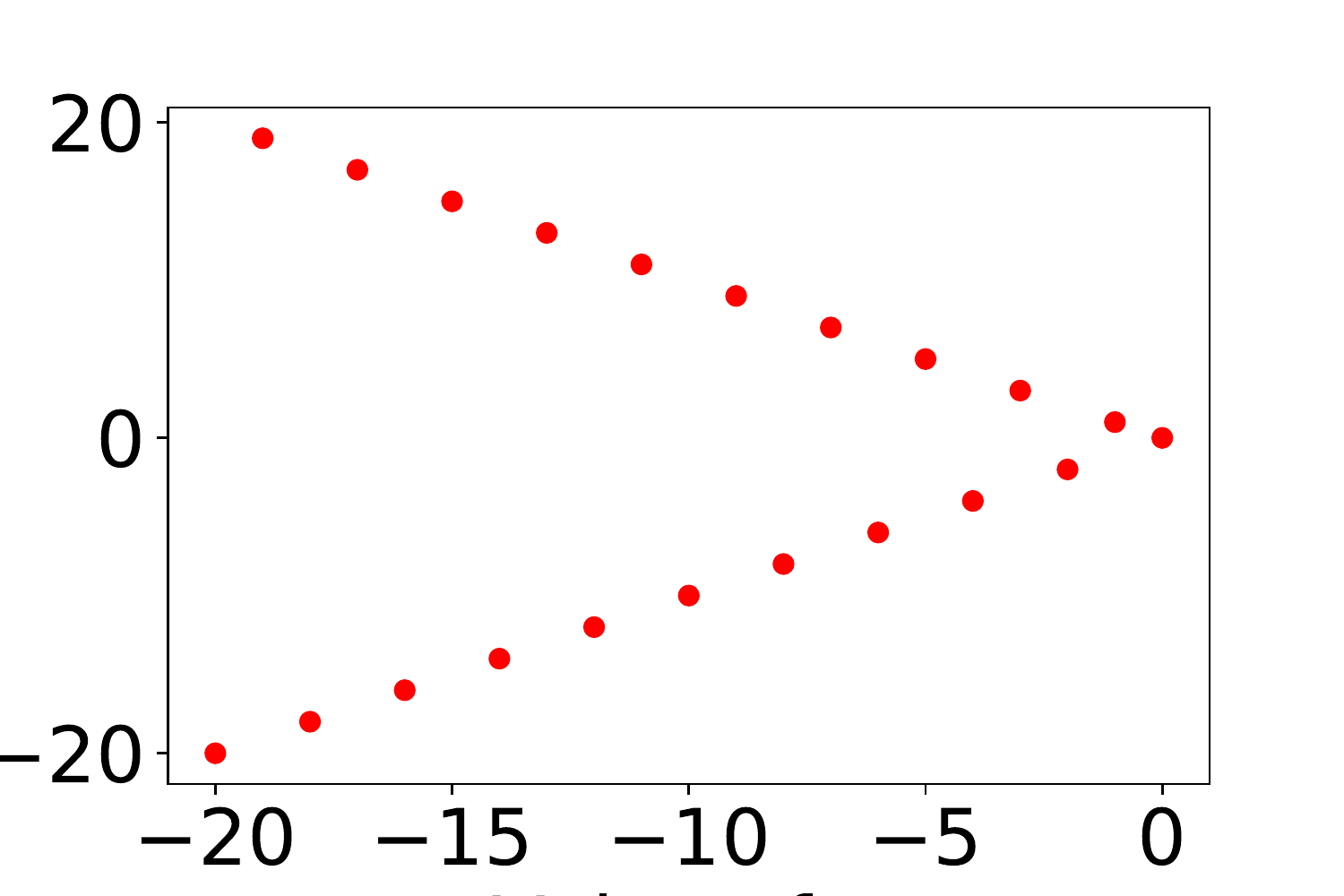}
\caption{\label{fig:plot_trace}Plotted trace of program}
\end{subfigure}
\caption{\label{fig:trace}Finding the Loop Invariant of Problem 1. }
\vspace{-10pt}
\end{figure}

\parheader{Description of Problems.} In this section, we discuss in detail two of the more difficult loop invariant problems we have created. 
Consider the first example loop in Fig.\ref{fig:trace}. For a loop invariant to be usable in verifying this loop, it must be valid for the precondition $t=-20, u=-20$, the recursion step representing loop execution when the loop condition is satisfied $(u\neq 0)$, and the post condition $t = 0$ when the loop condition is no longer satisfied $(u = 0)$.  The correct and precise invariant for the program is $((t+u=0) \lor (t-u=0))\land(t \leq 0)$. The plot of the trace in Fig.\ref{fig:plot_trace} shows that the points lie on one of two lines expressible as linear equality constraints. These constraints along with the inequality can be learned from the execution trace in under 20 seconds. The invariant that is inferred is $((t+u=0) \lor (t-u=0))\land(t \leq 0)$, which Z3 verifies is a sufficient loop invariant.

Figure~\ref{lst:source-1} shows the pseudocode for the second program. The correct and precise loop invariant is $(t+u=0) \land (v+w=0) \land (u+w \geq 0)$. Our CLN can learn the this invariant with the t-norm in under 20 seconds. Both Code2inv and LoopInvGen time out within one hour without finding a solution. 

\begin{figure}[h]
    \centering
\begin{lstlisting}[style=myC,numbers=none]
//pre: t=-10 /\ u=10 /\ v=-10 /\ w=10
while (u + w > 0) {
    if (unknown()) {
        t++; u--;
    } else {
        v++; w--;
    }
}
//post: t=w /\ u=v
\end{lstlisting}
\caption{Pseudocode for Problem 2, which involves a conjunction of equalities}
\label{lst:source-1}
\end{figure}

\section{T-norms and T-conorms}
\label{app:norms}
\begin{figure}[H]
  \centering
  \begin{subfigure}[H]{0.47\columnwidth}
    \includegraphics[width=\linewidth]{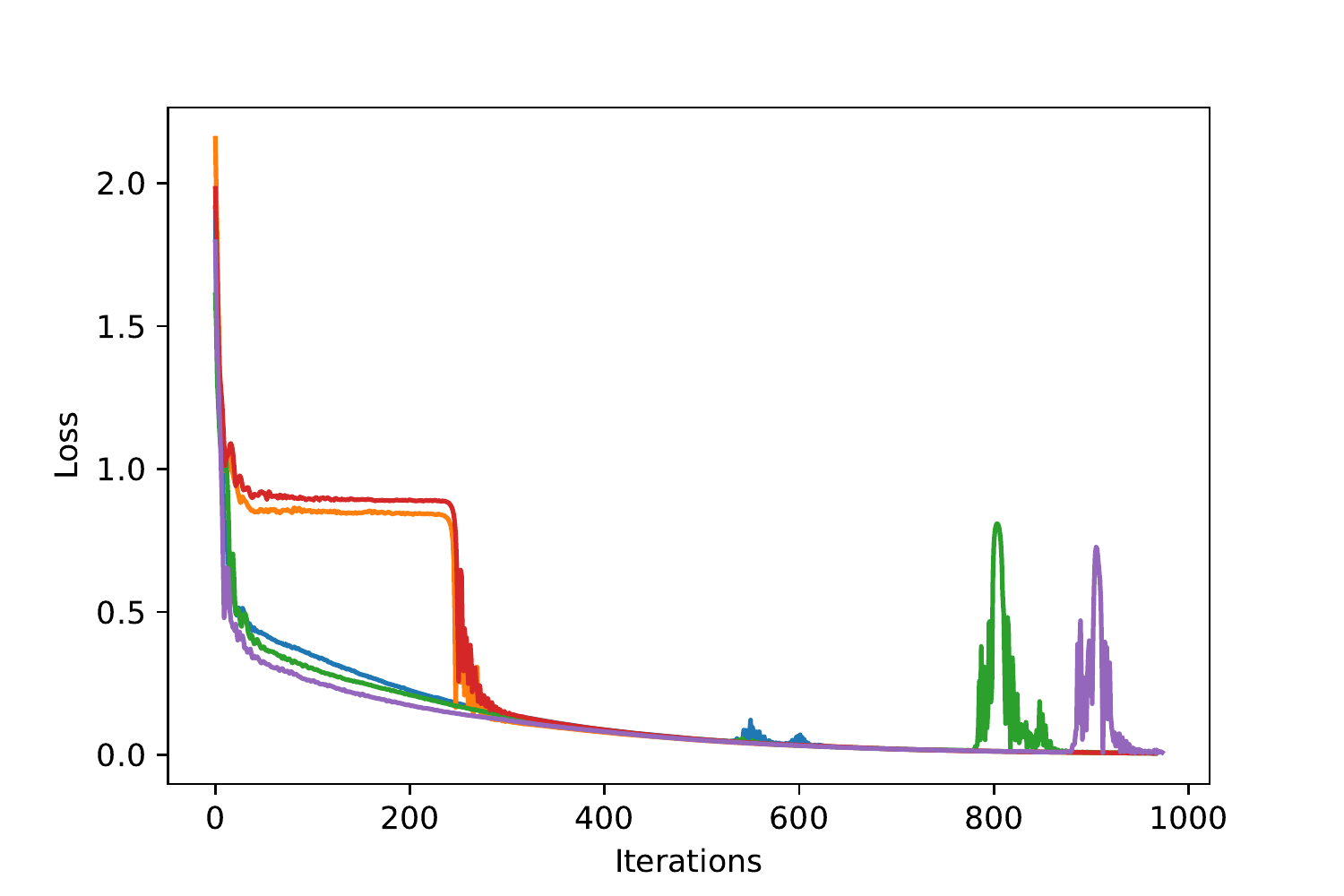}
    \caption{\label{fig:Godel-tnorm}{\rm Godel t-norm}}
  \end{subfigure}
  \begin{subfigure}[H]{0.47\columnwidth}
    \includegraphics[width=\linewidth]{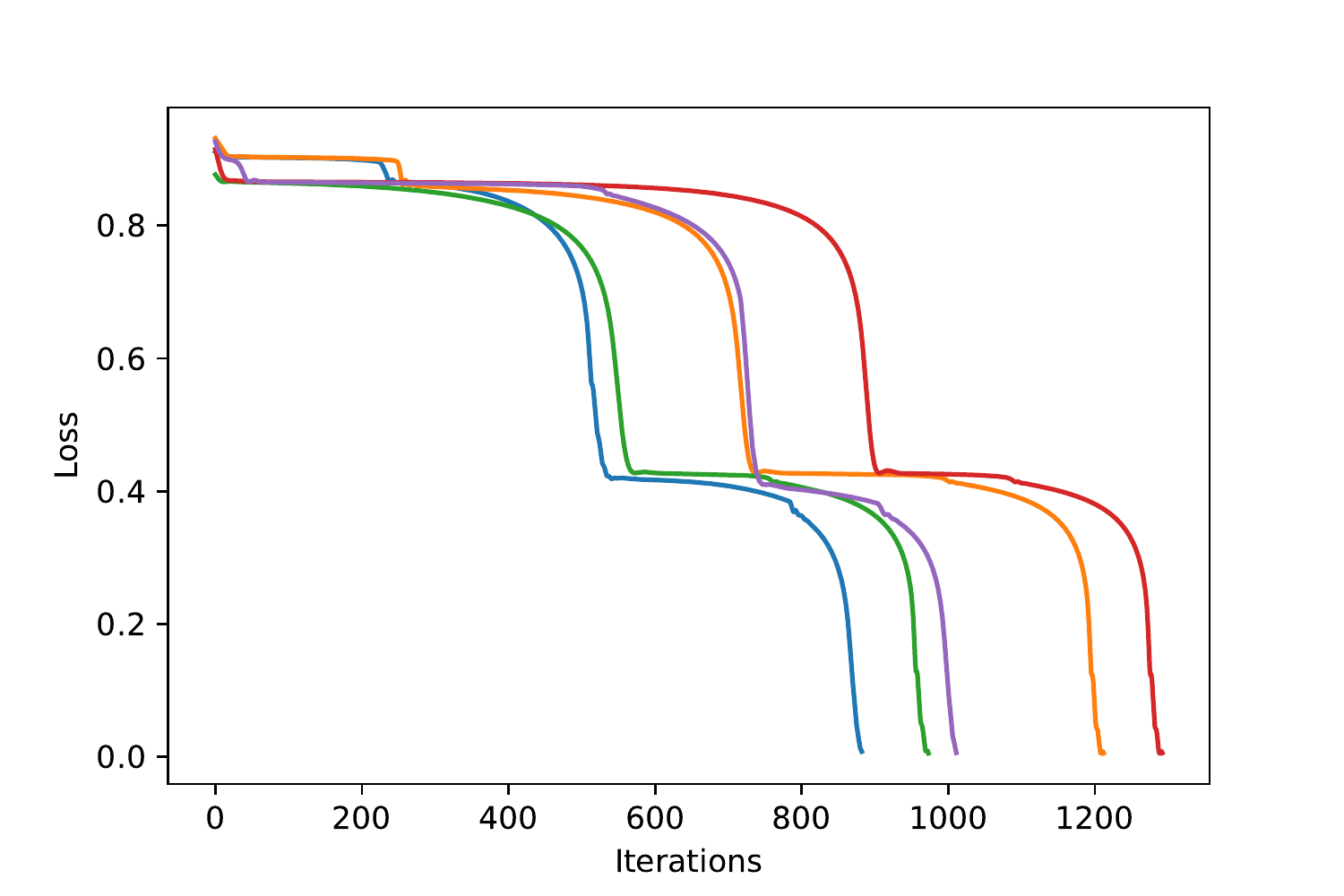}
    \caption{\label{fig:Godel-tconorm}{\rm Godel t-conorm}}
  \end{subfigure}
  \setlength{\belowcaptionskip}{-10pt}
\caption{\label{fig:godel_trace}Godel t-norm and t-conorm}
\end{figure}

\begin{figure}[H]
  \centering
  \begin{subfigure}[h]{0.47\columnwidth}
    \includegraphics[width=\linewidth]{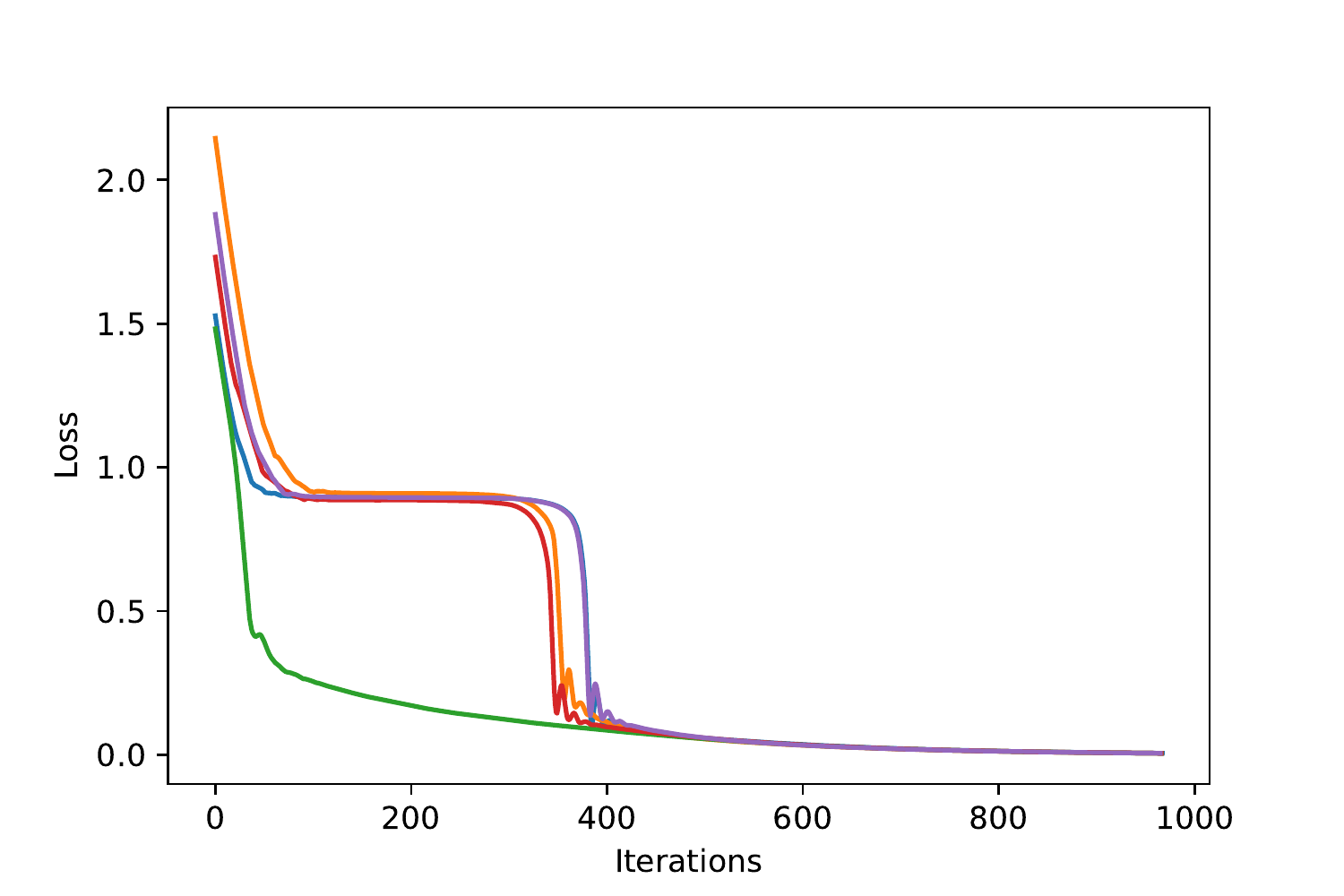}
    \caption{\label{fig:Luka-tnorm}{\rm Lukasiewicz t-norm}}
  \end{subfigure}
  \begin{subfigure}[h]{0.47\columnwidth}
    \includegraphics[width=\linewidth]{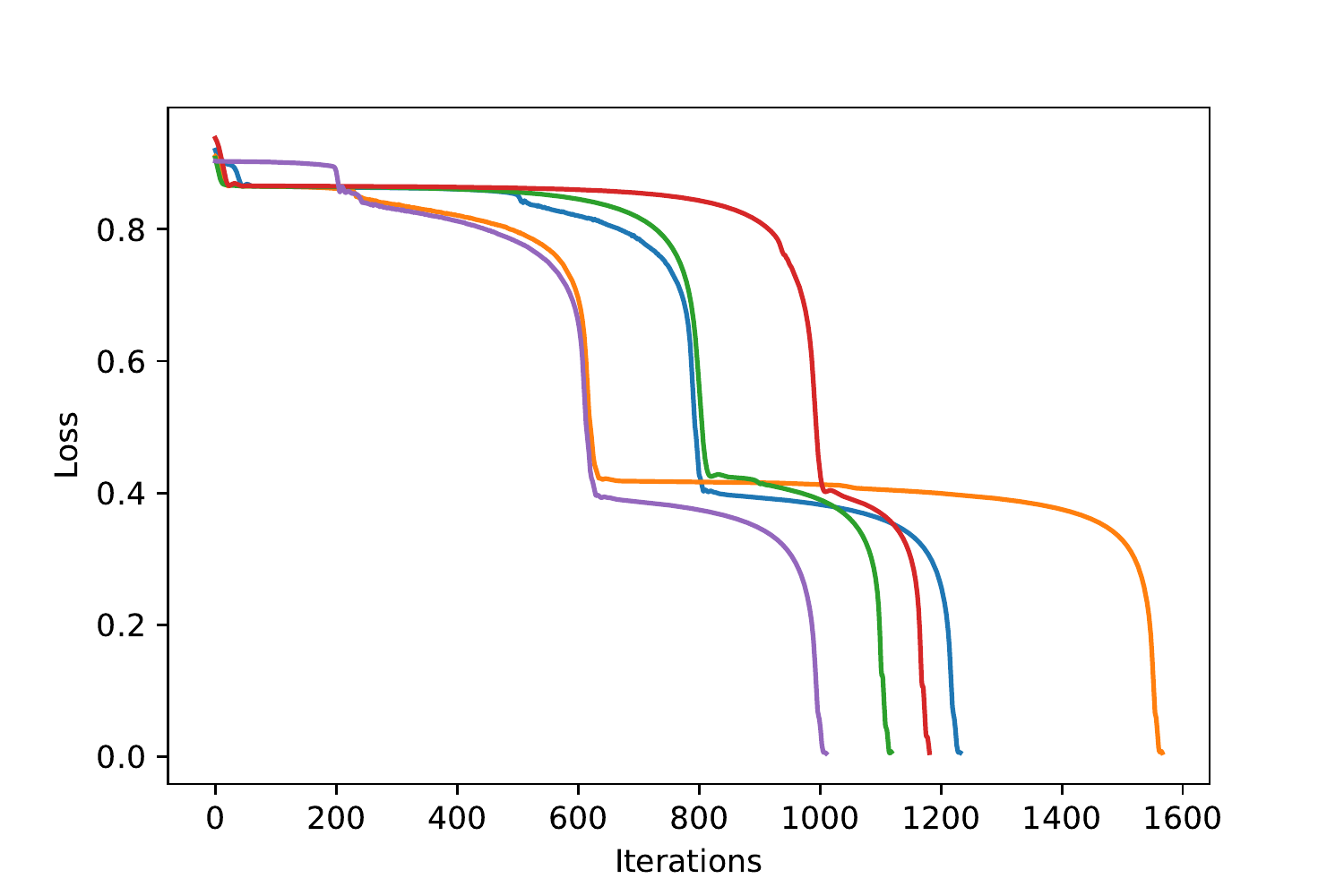}
    \caption{\label{fig:Luka-tconorm}{\rm Lukasiewicz t-conorm}}
  \end{subfigure}
  \setlength{\belowcaptionskip}{-10pt}
\caption{\label{fig:lukasiewicz_trace}Lukasiewicz t-norm and t-conorm}
\end{figure}

\begin{figure}[H]
  \centering
  \begin{subfigure}[H]{0.47\columnwidth}
    \includegraphics[width=\linewidth]{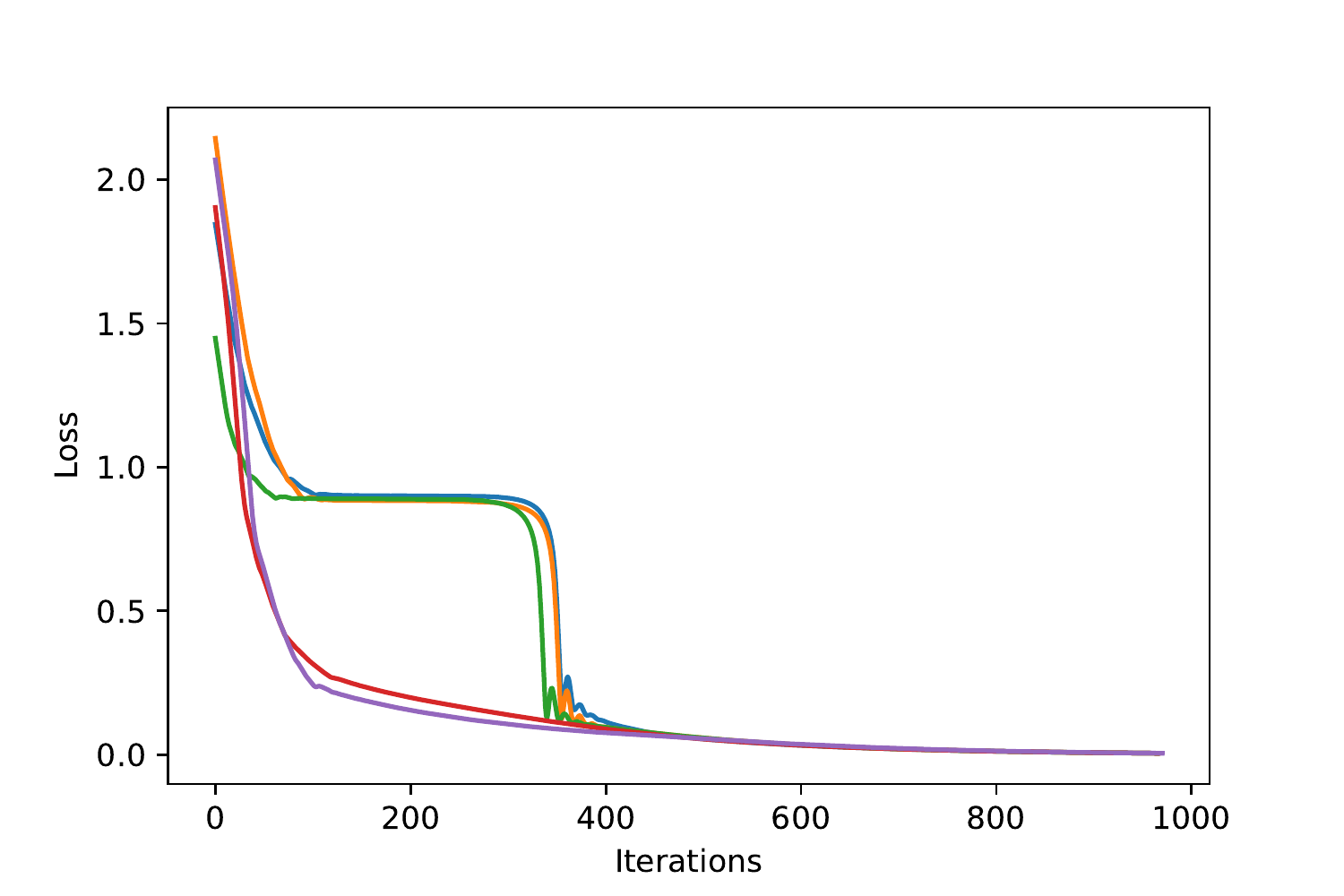}
    \caption{\label{fig:product-tnorm}{\rm Product t-norm}}
  \end{subfigure}
  \begin{subfigure}[H]{0.47\columnwidth}
    \includegraphics[width=\linewidth]{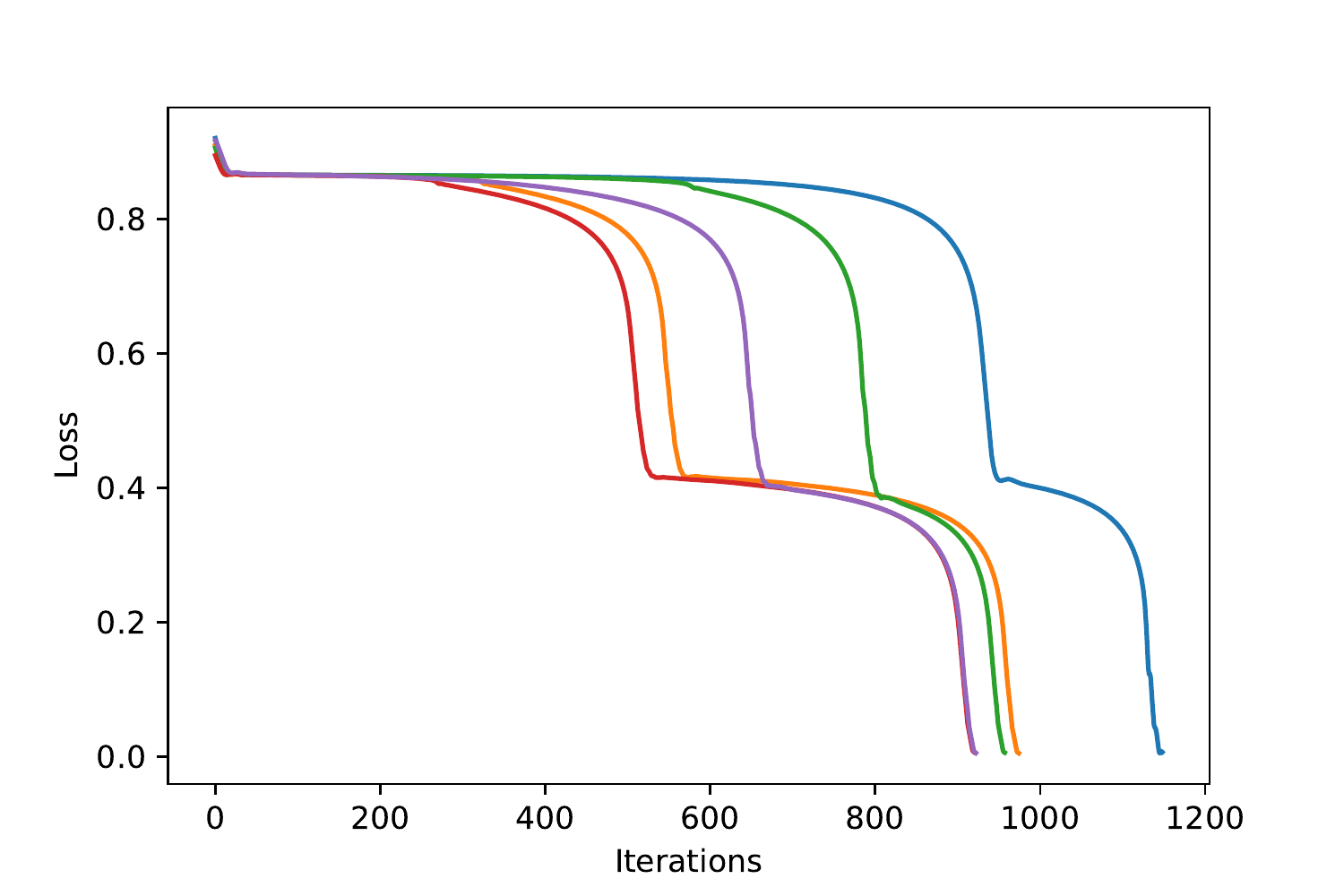}
    \caption{\label{fig:product-tconorm}{\rm Product t-conorm}}
  \end{subfigure}
  \setlength{\belowcaptionskip}{-10pt}
\caption{\label{fig:product_trace}Product t-norm and t-conorm}
\end{figure}

Here we provide more details on the convergence of t-norms and T-conorms on our models. 
 Figures \ref{fig:godel_trace},  \ref{fig:lukasiewicz_trace}, and \ref{fig:product_trace} show plots of the  training loss for Godel, Lukasiewicz,  and product t-norms and t-conorms respectively. Variations between individual traces are a result of random parameter initialization. CLNs with t-norms usually converge rapidly but sometimes will temporarily plateau when one clause converges before the other. In contrast, t-conorms always train one clause at a time, resulting in a curved staircase shape in the loss traces. We observe no significant difference between the t-norms used for conjunctions but observe that the product co-tnorms converges faster than the other t-conorms.

\end{appendices}

\end{document}